\documentclass{article}

\usepackage[utf8]{inputenc}
\usepackage{times}
\usepackage{graphicx}
\usepackage{subfigure}
\usepackage{natbib}
\usepackage{algorithm}
\usepackage{algorithmic}
\usepackage{hyperref}
\usepackage{pgfplots}
\usepackage{sansmath}
\usepackage{helvet}
\usepackage{bm}
\usepackage{amsmath}
\usepackage{amssymb}
\usepackage[accepted]{icml2015}

\usepgfplotslibrary{external}

\icmltitlerunning{A trust-region method for stochastic variational inference}

\DeclareMathSizes{10}{9}{7}{6}

\begin{document}

\twocolumn[
\icmltitle{A trust-region method for stochastic variational inference \\
with applications to streaming data}

\icmlauthor{Lucas Theis}{lucas@bethgelab.org}
\icmladdress{Centre for Integrative Neuroscience,
           Otfried-Müller-Str. 25, BW 72076 Germany}
\icmlauthor{Matthew D. Hoffman}{mathoffm@adobe.com}
\icmladdress{Adobe Research,
            601 Townsend St., San Francisco, CA 94103 USA}
\icmlkeywords{stochastic variational inference, latent dirichlet allocation, streaming,
variational Bayes, empirical Bayes, machine learning, ICML}

\vskip 0.3in
]

\begin{abstract}
	Stochastic variational inference allows for fast posterior inference in complex Bayesian models.
	However, the algorithm is prone to local optima which can make the quality of the posterior approximation
	sensitive to the choice of hyperparameters and initialization. We address this problem by
	replacing the natural gradient step of stochastic varitional inference with a
	trust-region update. We show that this leads to generally better results and reduced sensitivity to
	hyperparameters. We also describe a new strategy for variational inference on streaming data and show
	that here our trust-region method is crucial for getting good performance.
\end{abstract}

\section{Introduction}
Stochastic variational inference \citep[SVI;][]{Hoffman:2013} has enabled variational inference on
massive datasets for a large class of complex Bayesian models. It has been applied to, for example, topic models
\cite{Hoffman:2010}, nonparametric models \cite{Wang:2011,Paisley:2012}, mixture models \cite{Hughes:2013}, and matrix
factorizations \cite{Gopalan:2014}.
However, it has been observed that SVI can be sensitive to the choice of hyperparameters and is
prone to local optima \cite{Ranganath:2013,Hughes:2013,Hoffman:2015}. Successful attempts to improve
its performance include the automatic tuning of learning rates \cite{Ranganath:2013} and variance reduction
techniques \cite{Mandt:2014}. The results of \citet{Hoffman:2015} suggest that local optima in the objective function caused by
a mean-field approximation \cite{Wainwright:2008} can be eliminated by means of more complex families of approximating distributions.
They also find that the standard search heuristics used in variational inference consistently fail to find the best local optima
of the mean-field objective function,
suggesting that variational inference algorithms could be improved by employing more robust optimization algorithms.
In the first part of this paper we propose a new learning algorithm 
that replaces the natural gradient steps at the core of SVI with trust-region updates.
We show that these trust-region updates are feasible in practice and lead to a more robust algorithm generally yielding better performance.

In the second part of the paper we study the setting of continuous streams of data. SVI is a promising candidate for fitting Bayesian models
to data streams, yet the dependence of its updates on the dataset size and its underlying assumption of uniform sampling from the
entire dataset have hampered its application in practice. We therefore propose a new strategy for applying SVI
in the streaming setting. We show that this strategy fails when used with natural gradient steps but works well with our trust-region method.
When applied to latent Dirichlet allocation \citep[LDA;][]{Blei:2003}, our method is able to continuously integrate new data points
not only at the document level but also at the word level, which existing methods for variational inference on streaming data cannot do.

\section{A trust-region method for stochastic variational inference}
	In the following section, after introducing some notation and outlining the model class studied in this paper,
	we briefly review stochastic variational inference (SVI) in order to facilitate comparison with our trust-region extension.

	\subsection{Model assumptions}
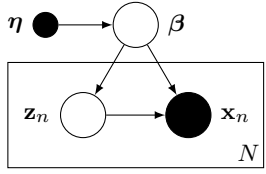
\begin{figure}[t]
			\centering
			\begin{tikzpicture}[
					unit/.style={draw,circle,minimum size=.6cm},
					param/.style={draw,circle,minimum size=.1cm,fill=black}
				]
				\node (e) at (-1.2, 1.2) [param,label=left:$\bm{\eta}$] {};
				\node (B) at (0.0, 1.2) [unit,label=right:$\bm{\beta}$]  {};
				\node (z) at (-.7, 0.0) [unit,label=left:$\mathbf{z}_n$] {};
				\node (x) at (0.7, 0.0) [unit,label=right:$\mathbf{x}_n$,fill=black] {};
				\draw[-latex] (z) -- (x);
				\draw[-latex] (B) -- (x);
				\draw[-latex] (B) -- (z);
				\draw[-latex] (e) -- (B);
				\draw (-1.7, -.7) rectangle (1.7, .7);
				\node at (1.5, -.5) {$N$};
			\end{tikzpicture}
			\caption{A graphical model representation of the model class considered in this paper.}
			\label{fig:model}
		\end{figure}

		Our basic model assumptions are summarized by\footnotemark
		\begin{align}
			\textstyle
			p(\bm{\beta}) &= h(\bm{\beta}) \exp\left( \bm{\eta}^\top t(\bm{\beta}) - a(\bm{\eta}) \right),
		\end{align}
		\begin{align}
			\textstyle
			p(\mathbf{x}, \mathbf{z} \mid \bm{\beta}) &= \prod_n h(\mathbf{x}_n,
			\mathbf{z}_n) \exp\left( t(\bm{\beta})^\top f(\mathbf{x}_n, \mathbf{z}_n)\right).
		\end{align}
		That is, we have \textit{global parameters} $\bm{\beta}$ whose prior distribution is an exponential family
		governed by natural parameters $\bm{\eta}$, and $N$ conditionally independent pairs of \textit{local parameters} $\mathbf{z}_n$
		and observations $\mathbf{x}_n$ whose exponential-family distribution is controlled by $\bm{\beta}$ (Figure~\ref{fig:model}).
		 While the basic strategy presented in this paper might
		also be applied in the non-conjugate case, for simplicity we assume
		conjugacy between the two exponential-family distributions.
		\footnotetext{The log-normalizer of the likelihood $p(\mathbf{x}, \mathbf{z} \mid \bm{\beta})$ is absorbed into $t(\bm{\beta})$ and $f(\mathbf{x}_n, \mathbf{z}_n)$ to simplify the notation.}

		Instances of this model class considered in greater detail in this paper are latent Dirichlet allocation
		\cite{Blei:2003} and mixture models. Other instances include (but are not limited to) hidden Markov models,
		probabilistic matrix factorizations, and hierarchical linear and probit regression \cite{Hoffman:2015}.

	\subsection{Mean-field variational inference}
		Mean-field variational inference approximates the posterior distribution over
		latent variables with a factorial distribution, which here we take to be of the form
		\begin{align}
			\textstyle
			q(\bm{\beta}, \mathbf{z}) &= q(\bm{\beta}) \prod_{n,m} q(z_{nm}), \\
			q(\bm{\beta}) &= h(\bm{\beta}) \exp\left( \bm{\lambda}^\top t(\bm{\beta}) - a(\bm{\lambda}) \right).
		\end{align}
		We further assume that $q(z_{nm})$ is controlled by parameters $\bm{\phi}_{nm}$. 
		Inference proceeds by alternatingly updating each factor to maximize
		the \textit{evidence lower bound} (ELBO),
		\begin{align}
			\label{eq:lower_bound1}
			\hspace{-0.2cm}
			\textstyle
			\mathcal{L}(\bm{\lambda}, \bm{\phi})
			&= \mathbb{E}_q\left[ \log \frac{p(\bm{\beta})}{q(\bm{\beta})} \right]
				+ \sum_{n = 1}^N \mathbb{E}_q\left[ \log \frac{p(\mathbf{x}_n, \mathbf{z}_n \mid \bm{\beta})}{q(\mathbf{z}_n)} \right],
		\end{align}
		which is equivalent to minimizing the Kullback-Leibler divergence between $q$
		and the true posterior distribution over $\bm{\beta}$ and $\mathbf{z}$.
		While the trust-region method described below might also be applied to more complex approximations, we are
		particularly interested to see how much a mean-field approximation can be improved via better learning algorithms.
		As mentioned above, the results of \citet{Hoffman:2015} suggest that mean-field approximations might often perform poorly
		due to local optima rather than the inflexibility of the approximating family \citep[see also][]{Wainwright:2008}. We therefore
		focus our attention on the mean-field approximation in the following.

	\subsection{Stochastic variational inference}
		Extending the work by \citet{Sato:2001}, \citet{Hoffman:2013} have shown that mean-field variational inference is equivalent to
		natural gradient ascent on
		\begin{align}
			\textstyle
			\label{eq:lower_bound2}
			\mathcal{L}(\bm{\lambda}) = \max_{\bm{\phi}} \mathcal{L}(\bm{\lambda}, \bm{\phi}).
		\end{align}
		This interpretation enables the derivation of stochastic natural gradient algorithms for variational inference,
		in this context called \textit{stochastic variational inference} (SVI).

		For a uniformly at random selected data point $n$,
		\begin{align}
			\hspace{-0.16cm}
			\mathcal{L}_n(\bm{\lambda}, \bm{\phi}_n)
			&= N \mathbb{E}_{q}\left[ \log \frac{p(\mathbf{x}_n, \mathbf{z}_n \mid \bm{\beta})}{q(\mathbf{z}_n)} \right]
			+ \mathbb{E}_{q}\left[ \log \frac{p(\bm{\beta})}{q(\bm{\beta})} \right]
		\end{align}
		represents an unbiased stochastic approximation of the ELBO given in Equation~\ref{eq:lower_bound1}.
		Similarly, 
		\begin{align}
			\textstyle
			\mathcal{L}_n(\bm{\lambda}) &= \mathcal{L}_n(\bm{\lambda}, \bm{\phi}_n^*), \\
			\label{eq:phi_update}
			\bm{\phi}_n^* &= \text{argmax}_{\bm{\phi}_n}\, \mathcal{L}_n(\bm{\lambda}, \bm{\phi}_n)
		\end{align}
		represents an unbiased estimate of the lower bound in Equation~\ref{eq:lower_bound2}. For simplicity,
		in the following we only consider stochastic approximations based on a single data point. An extension to batches
		of multiple data points is straightforward.
		A step in the direction of the natural gradient of $\mathcal{L}_n(\bm{\lambda})$ scaled by a learning rate
		of $\rho_t$ is given by \cite{Hoffman:2013}
		\begin{align}
			\label{eq:ng}
			\bm{\lambda}_{t + 1}
			&= (1 - \rho_t) \bm{\lambda}_t + \rho_t \left( \bm{\eta} + N \mathbb{E}_{\bm{\phi}_n^*}[f(\mathbf{x}_n, \mathbf{z}_n)] \right).
		\end{align}

	\subsection{Trust-region updates}
		As has been observed before \cite{Hughes:2013,Hoffman:2015} and as we will corroborate further in Section~\ref{sec:experiments}, 
		SVI can be prone to local optima. One possible solution to this problem is to use a more flexible family of approximating
		distributions \cite{Hoffman:2015}, effectively smoothing the objective function but sacrificing the speed and convenience
		of a mean-field approximation. In contrast, here we try to address the issue of local optima by improving on the
		optimization algorithm.

		Instead of performing a natural gradient step (Equation~\ref{eq:ng}), we propose the following stochastic update:
		\begin{align}
			\textstyle
			\label{eq:tr}
			\bm{\lambda}_{t + 1} = \text{argmax}_{\bm{\lambda}}\, \left\{ \mathcal{L}_n(\bm{\lambda})
			- \xi_t D_\text{KL}(\bm{\lambda}, \bm{\lambda}_t) \right\}.
		\end{align}
		Intuitively, this \textit{trust-region step} tries to find the optimal solution to a
		stochastic approximation of the lower bound, regularized to prevent the distribution over parameters from changing too much.
                The degree of change between distributions is measured by the Kullback-Leibler (KL) divergence $D_\text{KL}$, and the parameter
                $\xi_t > 0$ controls the strength of the regularization.
		If we could solve Equation~\ref{eq:tr} in closed form, this algorithm would at least no longer be hampered by local optima
		in $\mathcal{L}_n$. In general this will not be possible so that we have to resort to the following optimization scheme.
		\begin{algorithm}[tb]
			\caption{Trust-region SVI}
			\label{alg:trsvi}
			\begin{algorithmic}
				\STATE Set $t=0$, initialize $\bm{\lambda}_0$ randomly
				\REPEAT
					\STATE Select $\mathbf{x}_n$ uniformly from the dataset
					\STATE Initialize $\bm{\phi}^*_n$
					\REPEAT
						\STATE $\textstyle\bm{\lambda} \leftarrow (1 - \rho_t) \bm{\lambda}_t + \rho_t \left( \bm{\eta} + N \mathbb{E}_{\bm{\phi}_n^*}[f(\mathbf{x}_n, \mathbf{z}_n)] \right)$
						\STATE $\textstyle\bm{\phi}_n^* \leftarrow \text{argmax}_{\bm{\phi}_n} \mathcal{L}_n(\bm{\lambda}, \bm{\phi}_n)$
					\UNTIL{convergence}
					\STATE $\bm{\lambda}_{t + 1} \leftarrow \bm{\lambda}$
                                        \STATE $t\leftarrow t+1$
				\UNTIL{convergence}
			\end{algorithmic}
		\end{algorithm}

		For fixed $\bm{\phi}_n^*$, the natural gradient of the objective on the right-hand side of Equation~\ref{eq:tr} is given by
		\begin{align}
			\label{eq:tr_gradient}
			\bm{\eta} + N \mathbb{E}_{\bm{\phi}_n^*}[f(\mathbf{x}_n, \mathbf{z}_n)] -
			\bm{\lambda} + \xi_t (\bm{\lambda}_t - \bm{\lambda}).
		\end{align}
		This follows from the the natural gradient of $\mathcal{L}_n$ \cite{Hoffman:2013} and the
		fact that the gradient of the Kullback-Leibler divergence between two distributions from an
		expnential family in canonical form is given by
		\begin{align}
			\nabla_{\bm{\lambda}} D_\text{KL}(\bm{\lambda}, \bm{\lambda}') = I(\bm{\lambda}) (\bm{\lambda} - \bm{\lambda}'),
		\end{align}
		where $I(\bm{\lambda})$ is the Fisher information matrix (see Supplementary Section~1 for a derivation).
		Setting the gradient to zero yields
		\begin{align}
			\label{eq:lambda_update}
			\bm{\lambda} = (1 - \rho_t) \bm{\lambda}_t + \rho_t \left( \bm{\eta} + N \mathbb{E}_{\bm{\phi}_n^*}[f(\mathbf{x}_n, \mathbf{z}_n)] \right),
		\end{align}
		defining $\rho_t \equiv (1 + \xi_t)^{-1}$. We propose to solve Equation~\ref{eq:tr} via alternating coordinate ascent, that is,
		by alternatingly computing $\bm{\phi}_n^*$ (Equation~\ref{eq:phi_update}) and $\bm{\lambda}$ (Equation~\ref{eq:lambda_update})
		until some stopping criterion is reached.
		Note that natural gradient ascent can be seen as a special case of this
		trust-region method where $\bm{\lambda}$ is initialized with $\bm{\lambda}_t$ and only one iteration of updates to
		$\bm{\phi}_n^*$ and $\bm{\lambda}$ is performed.

		The need to iteratively optimize $\bm{\lambda}$ makes each trust-region step more costly than a simple
		natural gradient step. However, the additional overhead is often smaller than one might
		expect. For many models (such as LDA), the dominant cost is solving the sub-problem
		of computing $\text{argmax}_{\phi_n} \mathcal{L}(\bm{\lambda}, \bm{\phi}_n^*)$,
		which must be done iteratively. This sub-problem can be initialized with the previous value
		of $\bm{\phi}_n^*$; when $\bm{\lambda}$ is near convergence, this initialization
		will already be near a (local) optimum, and the sub-problem can be solved quickly.

		In the limit of large $\xi_t$, the solution to Equation~\ref{eq:tr} will become identical to the natural gradient
		step in Equation~\ref{eq:ng} (see Supplementary Section~2). Consequently, the convergence guarantees that
		hold for SVI carry over to our trust region method. That is, under certain regularity assumptions and for appropriately
		decaying $\rho_t$ \cite{Bottou:1998}, iteratively applying Equation~\ref{eq:tr} will converge to a local optimum
		of $\mathcal{L}$.

		For any finite $\xi_t$, the two update steps will generally be different. A crucial advantage of the trust-region
		method is that in each iteration we can initialize $\bm{\lambda}$ and $\bm{\phi}_n^*$ arbitrarily before applying the alternating
		optimization scheme. This way we can hope to jump out of local optima. The optimal initialization will
		depend on the data and the specific model being used. However, in our experiments with LDA and mixture models we found that
		a generally useful strategy is to initialize $\bm{\phi}_n^*$ such that the beliefs over $\mathbf{z}_n$ are uniform and
		to compute the initial $\bm{\lambda}$ with these beliefs.

		The general algorithm is summarized in Algorithm~\ref{alg:trsvi}. More detailed derivations for LDA and mixture models
		can be found in the supplementary material.

\subsection{Related work}
	Our optimization resembles \textit{mirror descent} \cite{Nemirovski:1983,Beck:2003}, which
	applied to the lower bound (Equation~\ref{eq:lower_bound2}) corresponds to updates of the form
	\begin{align}
		\textstyle
		\label{eq:mirror_descent}
		\hspace{-.2cm}
		\bm{\lambda}_{t + 1} = \text{argmax}_{\bm{\lambda}}\, \left\{
			\left< \nabla_{\bm{\lambda}} \mathcal{L}_n(\bm{\lambda}_t), \bm{\lambda} - \bm{\lambda}_t \right>
			- \xi_t B(\bm{\lambda}, \bm{\lambda}_t) \right\}
	\end{align}
	for some Bregman divergence $B$. The main difference to our algorithm is that we try to optimize
	$\mathcal{L}_n$ exactly instead of a first-order approximation.

	Trust-region methods have a long history in optimization \cite{Nocedal:1999} and are frequently
	brought to bear on machine learning problems \citep[e.g.,][]{Lin:2007,Pascanu:2014}. However, we
	are unaware of any other trust-region method which omits a local approximation
	(Equation~\ref{eq:tr}).

\section{Streaming}
	Despite its sequential nature, SVI has not found widespread use in the streaming setting.
	One reason is that it assumes that the dataset is fixed and its size, $N$, known in advance. In contrast, 
	streaming data only becomes available over time and is potentially infinite.
	Another disadvantage of a naive application of SVI to the streaming setting where data points are processed once
	as they arrive is that it might not make the best use of the available computational resources. In real world
	applications, data points rarely arrive at a constant rate.
	Processing data points as they arrive thus means that there will be times when a computer has to quickly process many
	data points and other times where it will be idle.

	In the following, we propose an alternative but similarly straightforward application of SVI to the
	streaming setting. As we will show in Section~\ref{sec:experiments}, this algorithm gives poor results with natural gradient
	steps as local optima and sensitivity to hyperparameters are particularly problematic in the streaming setting. However,
	we find that it performs well using our trust-region updates.

	\subsection{Streaming SVI}
		\label{sec:streamingsvi}
		Rather than processing a data point once when it arrives, we suggest continuously
		optimizing an evolving lower bound. Instead of updating parameters directly, each new data
		point is simply added to a database. At time $t$, we optimize 
		\begin{align}
			\label{eq:lower_bound3}
			\hspace{-0.2cm}
\mathbb{E}_q\left[ \log \frac{p(\bm{\beta})}{q(\bm{\beta})} \right] +
			\max_{\bm{\phi}} \sum_{n = 1}^{N_t} \mathbb{E}_q\left[ \log \frac{p(\mathbf{x}_n, \mathbf{z}_n \mid \bm{\beta})}{q(\mathbf{z}_n)} \right],
		\end{align}
		where $N_t$ is the number of observed data points, by selecting a data point (or batch of data points) from
		the database uniformly at random and performing either a natural gradient or trust-region
		update. 

		Learning algorithms whose performance is robust to changes in hyperparameter settings are
		especially important in the streaming setting where performing test runs and
		cross-validation is often not an option. In addition to using trust-region updates
		we therefore employ empirical Bayes. After updating $\bm{\lambda}$ by performing a
		trust-region step (Equation~\ref{eq:tr}), we update $\bm{\eta}$ by performing one stochastic
		step in the direction of the natural gradient of the lower bound,
		\begin{align}
			\textstyle
			\label{eq:empirical_bayes}
			\bm{\eta}_{t + 1} = \bm{\eta}_t + \rho_t \left( E_{\bm{\lambda}_{t + 1}}\left[ t(\bm{\beta}) \right] - E_{\bm{\eta}_t}\left[ t(\bm{\beta}) \right] \right).
		\end{align}
		To make the learning algorithm and hyperparameters less dependent on the speed with which new data points
		arrive, we use the following learning rate schedule,
		\begin{align}
			\textstyle
			\rho_t = \left(\tau + \frac{N_t - N_0}{B}\right)^{-\kappa},
		\end{align}
		where $B$ is the batch size used. Instead of coupling the learning rate to the number of updates to the
		model, this schedule couples it to the number of data points added to the database since the start of learning,
		$N_t - N_0$.

	\subsection{Streaming variational Bayes}
		Streaming variational Bayes (SVB) has been proposed as an alternative to SVI 
		for the streaming setting \cite{Broderick:2013,Tank:2014}. Unlike SVI, its updates are independent of
		the dataset size. The basic idea behind SVB is that of assumed density filtering
		\citep[e.g.,][]{Maybeck:1982,Minka:2001}: the posterior is updated one data point at a time using Bayes rule and approximated between each two updates.
		Applied to our model class and ignoring the local variables for a moment, an update can be summarized as
		\begin{align}
			\label{eq:svb}
			\tilde p_{n + 1}(\bm{\beta}) &\propto p(\mathbf{x}_n \mid \bm{\beta}) q_{\bm{\lambda}_n}(\bm{\theta}), \\
			\bm{\lambda}_{n + 1} &= \text{argmin}_{\bm{\lambda}}\, D_\text{KL}\left[ q_{\bm{\lambda}}(\bm{\beta}) \mid\mid \tilde p_{n + 1}(\bm{\beta}) \right],
		\end{align}
		where $\bm{\lambda}_0$ is set to $\bm{\eta}$.


	\subsection{Streaming batch algorithm}
		\label{sec:streamingbatch}
		We will compare our streaming SVI algorithm and SVB to the following simple baseline algorithm.
		At each iteration, we randomly select a large batch of $\min(B, N_t)$ data points from the database, where
		$N_t$ is the current number of data points in the database. We then run a batch
		algorithm to perform mean-field variational inference for a fixed number of iterations (or until it
		converges). Once the training is complete, we replace the current parameters with the parameters found by the batch
		algorithm and immediately restart the training on a newly sampled dataset of size $B$.

\section{Experiments}
	\label{sec:experiments}
	In the following we demonstrate the usefulness of our proposed modifications to SVI. We start with a toy example
	and continue with increasingly realistic applications.

	\subsection{MNIST}
		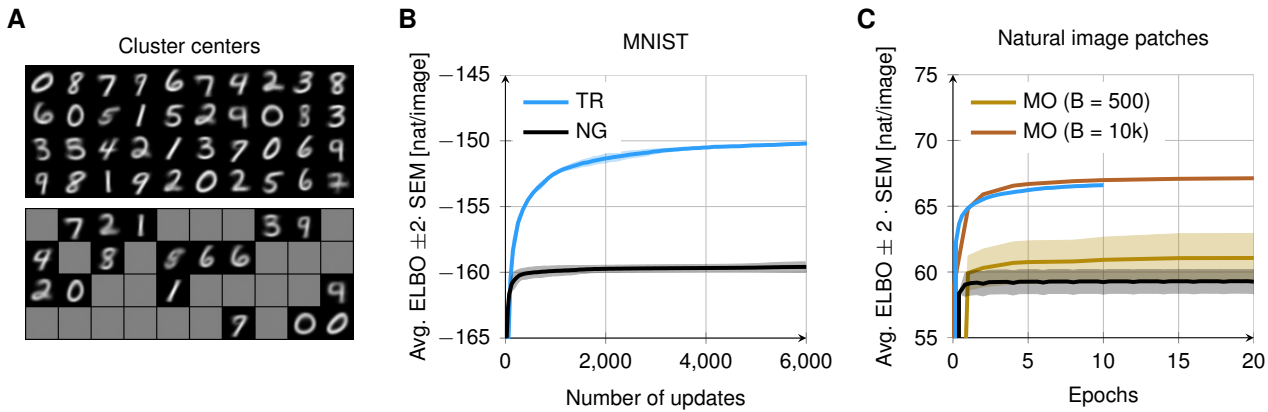
\begin{figure*}[t]
		\begin{tikzpicture}
			\node at (-6.8cm, 4.25cm) {\sffamily\textbf A};
			\node at (-1.6cm, 4.25cm) {\sffamily\textbf B};
			\node at (4.5cm, 4.25cm) {\sffamily\textbf C};
			\node at (-4.5cm, 3.9cm) {\fontsize{8pt}{8pt}\sffamily\selectfont Cluster centers};
			\node at (-4.5cm, 2.75cm) {\includegraphics[height=1.75cm]{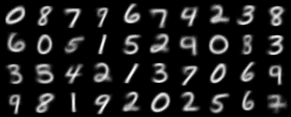}};
			\node at (-4.5cm, 0.85cm) {\includegraphics[height=1.75cm]{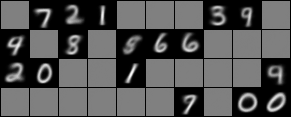}};
			\begin{axis}[
				scale only axis,
				width=4cm,
				height=3.5cm,
				at={(-0.3cm, 0.0cm)},
				title={MNIST},
				xmin=0,
				xmax=6000,
				ymin=-165,
				ymax=-145,
				xlabel={Number of updates},
				ylabel={Avg. ELBO $\pm 2 \cdot$ SEM [nat/image]},
				legend entries={TR,NG},
				legend cell align=left,
				legend style={
					at={(0.025,0.975)},
					anchor=north west,
					font=\fontsize{8pt}{8pt}\sansmath\sffamily\selectfont,
					draw=none
				},
				xlabel near ticks,
				ylabel near ticks,
				ylabel style={at={(-0.22, 0.5)}},
				grid=major,
				xtick={0,2000,4000,6000},
				xticklabel={\pgfmathprintnumber[precision=4]{\tick}},
				yticklabel={\pgfmathprintnumber[precision=4]{\tick}},
				zticklabel={\pgfmathprintnumber[precision=4]{\tick}},
				title style={font=\fontsize{8pt}{8pt}\sansmath\sffamily\selectfont},
				label style={font=\fontsize{8pt}{8pt}\sansmath\sffamily\selectfont},
				tick label style={font=\fontsize{8pt}{8pt}\sansmath\sffamily\selectfont},
				axis x line=bottom,
				axis y line=left]
				\addplot+[
					no marks,
					line width=0pt,
					color={rgb,255:red,45;green,158;blue,250},
					fill={rgb,255:red,45;green,158;blue,250},
					opacity=0.3,
					mark options={solid, fill opacity=0.3},
					forget plot] coordinates {
					(50.0, -165.34715044)
					(150.0, -157.464480675)
					(250.0, -155.820029045)
					(350.0, -155.036742382)
					(450.0, -154.392832793)
					(550.0, -153.878079857)
					(650.0, -153.333827518)
					(750.0, -153.001804853)
					(850.0, -152.797832356)
					(950.0, -152.443025143)
					(1050.0, -152.18099364)
					(1150.0, -152.028078717)
					(1250.0, -151.857465189)
					(1350.0, -151.694451445)
					(1450.0, -151.573835945)
					(1550.0, -151.437174341)
					(1650.0, -151.347479028)
					(1750.0, -151.209144553)
					(1850.0, -151.157415484)
					(1950.0, -151.060896552)
					(2050.0, -150.977963937)
					(2150.0, -150.915558435)
					(2250.0, -150.868319035)
					(2350.0, -150.844585144)
					(2450.0, -150.77755333)
					(2550.0, -150.759797251)
					(2650.0, -150.73135992)
					(2750.0, -150.687335415)
					(2850.0, -150.643285982)
					(2950.0, -150.637613542)
					(3050.0, -150.609933813)
					(3150.0, -150.60652617)
					(3250.0, -150.570252008)
					(3350.0, -150.570809015)
					(3450.0, -150.539894891)
					(3550.0, -150.527351642)
					(3650.0, -150.526475806)
					(3750.0, -150.486811469)
					(3850.0, -150.489800001)
					(3950.0, -150.445658655)
					(4050.0, -150.453241177)
					(4150.0, -150.438967384)
					(4250.0, -150.430312437)
					(4350.0, -150.419447621)
					(4450.0, -150.400997006)
					(4550.0, -150.40034257)
					(4650.0, -150.390478045)
					(4750.0, -150.386298364)
					(4850.0, -150.346952548)
					(4950.0, -150.32259696)
					(5050.0, -150.313413372)
					(5150.0, -150.290853456)
					(5250.0, -150.295260762)
					(5350.0, -150.28872911)
					(5450.0, -150.268842689)
					(5550.0, -150.250714742)
					(5650.0, -150.215239999)
					(5750.0, -150.19703038)
					(5850.0, -150.18186399)
					(5950.0, -150.189400176)
					(6050.0, -150.171686789)
					(6150.0, -150.170026678)
					(6250.0, -150.153674568)
					(6350.0, -150.124835296)
					(6450.0, -150.129575861)
					(6550.0, -150.092648708)
					(6650.0, -150.0934474)
					(6750.0, -150.092405784)
					(6850.0, -150.103069329)
					(6950.0, -150.091550568)
					(7050.0, -150.102468342)
					(7150.0, -150.052361547)
					(7250.0, -150.06392107)
					(7350.0, -150.051939081)
					(7450.0, -150.037130339)
					(7550.0, -150.049065819)
					(7650.0, -150.076932235)
					(7750.0, -150.047138797)
					(7850.0, -150.024584921)
					(7950.0, -150.014925882)
					(8050.0, -150.017488105)
					(8150.0, -150.029414121)
					(8250.0, -150.020731838)
					(8350.0, -150.013256699)
					(8450.0, -149.977743237)
					(8550.0, -149.979661965)
					(8650.0, -149.97561505)
					(8750.0, -149.977745333)
					(8850.0, -149.986653077)
					(8950.0, -149.966253156)
					(9050.0, -149.963635347)
					(9150.0, -149.951825715)
					(9250.0, -149.954307124)
					(9350.0, -149.929824078)
					(9450.0, -149.939245134)
					(9550.0, -149.918339821)
					(9650.0, -149.917340369)
					(9750.0, -149.922229235)
					(9850.0, -149.923670368)
					(9950.0, -149.921521557)
					(10050.0, -149.922399732)
					(10150.0, -149.888285332)
					(10250.0, -149.891216437)
					(10350.0, -149.891276134)
					(10450.0, -149.882363801)
					(10550.0, -149.886394342)
					(10650.0, -149.892803189)
					(10750.0, -149.8872142)
					(10850.0, -149.887311929)
					(10950.0, -149.880735718)
					(11050.0, -149.886130176)
					(11150.0, -149.884975871)
					(11250.0, -149.875959871)
					(11350.0, -149.878723132)
					(11450.0, -149.856395347)
					(11550.0, -149.854355763)
					(11650.0, -149.85475064)
					(11750.0, -149.86275594)
					(11850.0, -149.873452483)
					(11950.0, -149.847468775)
					(12000.0, -149.879080333)
					(11900.0, -149.886698124)
					(11800.0, -149.892256841)
					(11700.0, -149.896680152)
					(11600.0, -149.899224849)
					(11500.0, -149.914879866)
					(11400.0, -149.918309721)
					(11300.0, -149.933529005)
					(11200.0, -149.912592753)
					(11100.0, -149.921484811)
					(11000.0, -149.928722754)
					(10900.0, -149.924453005)
					(10800.0, -149.932106314)
					(10700.0, -149.946106154)
					(10600.0, -149.951721203)
					(10500.0, -149.953922955)
					(10400.0, -149.966826347)
					(10300.0, -149.957420462)
					(10200.0, -149.963251189)
					(10100.0, -149.96551019)
					(10000.0, -149.960460573)
					(9900.0, -149.953148747)
					(9800.0, -149.969778794)
					(9700.0, -149.982606834)
					(9600.0, -149.988087034)
					(9500.0, -149.991249599)
					(9400.0, -149.999802818)
					(9300.0, -149.9897285)
					(9200.0, -149.985596424)
					(9100.0, -149.988075542)
					(9000.0, -149.990703906)
					(8900.0, -150.004191192)
					(8800.0, -150.023897318)
					(8700.0, -150.019596203)
					(8600.0, -150.021275196)
					(8500.0, -150.043270717)
					(8400.0, -150.049987402)
					(8300.0, -150.061248614)
					(8200.0, -150.047324844)
					(8100.0, -150.060618555)
					(8000.0, -150.06791572)
					(7900.0, -150.066910723)
					(7800.0, -150.074658069)
					(7700.0, -150.08583523)
					(7600.0, -150.090941563)
					(7500.0, -150.09926477)
					(7400.0, -150.129836)
					(7300.0, -150.114854654)
					(7200.0, -150.127038105)
					(7100.0, -150.13436742)
					(7000.0, -150.123513083)
					(6900.0, -150.140737721)
					(6800.0, -150.137312593)
					(6700.0, -150.156576038)
					(6600.0, -150.16742984)
					(6500.0, -150.191029265)
					(6400.0, -150.198457016)
					(6300.0, -150.196054405)
					(6200.0, -150.191287607)
					(6100.0, -150.212145262)
					(6000.0, -150.230904424)
					(5900.0, -150.267667876)
					(5800.0, -150.271667216)
					(5700.0, -150.288084668)
					(5600.0, -150.303154303)
					(5500.0, -150.299943561)
					(5400.0, -150.317949101)
					(5300.0, -150.348480773)
					(5200.0, -150.371892832)
					(5100.0, -150.3664717)
					(5000.0, -150.367871803)
					(4900.0, -150.389262111)
					(4800.0, -150.392857438)
					(4700.0, -150.439103285)
					(4600.0, -150.436231659)
					(4500.0, -150.445547912)
					(4400.0, -150.495212407)
					(4300.0, -150.472562985)
					(4200.0, -150.485642866)
					(4100.0, -150.493410665)
					(4000.0, -150.562444606)
					(3900.0, -150.62126907)
					(3800.0, -150.627439485)
					(3700.0, -150.633426513)
					(3600.0, -150.664393866)
					(3500.0, -150.734876766)
					(3400.0, -150.741619953)
					(3300.0, -150.829877942)
					(3200.0, -150.84530632)
					(3100.0, -150.909705972)
					(3000.0, -150.976264964)
					(2900.0, -151.050653559)
					(2800.0, -151.152683819)
					(2700.0, -151.20416765)
					(2600.0, -151.256316022)
					(2500.0, -151.307714448)
					(2400.0, -151.376298931)
					(2300.0, -151.478670766)
					(2200.0, -151.560645339)
					(2100.0, -151.585286563)
					(2000.0, -151.673517075)
					(1900.0, -151.712708159)
					(1800.0, -151.770690966)
					(1700.0, -151.871128361)
					(1600.0, -151.938978534)
					(1500.0, -151.999699664)
					(1400.0, -152.104679361)
					(1300.0, -152.185100507)
					(1200.0, -152.294819017)
					(1100.0, -152.361053467)
					(1000.0, -152.646760034)
					(900.0, -152.770628816)
					(800.0, -153.224970084)
					(700.0, -153.664676659)
					(600.0, -153.848026365)
					(500.0, -154.255110543)
					(400.0, -154.971165176)
					(300.0, -155.97760499)
					(200.0, -157.718307096)
					(150.0, -159.213212459)
					(100.0, -161.466801605)
					(50.0, -167.700257609)
				};
				\addplot+[
					no marks,
					line width=1.5pt,
					color={rgb,255:red,45;green,158;blue,250},
					mark options={solid}] coordinates {
					(50.0, -166.523704025)
					(100.0, -160.552770721)
					(150.0, -158.338846567)
					(250.0, -156.262810795)
					(350.0, -155.220312659)
					(450.0, -154.492094424)
					(550.0, -153.964338183)
					(650.0, -153.566952605)
					(750.0, -153.233741261)
					(850.0, -152.853952676)
					(950.0, -152.555264591)
					(1050.0, -152.360101453)
					(1150.0, -152.170047943)
					(1250.0, -152.046685792)
					(1350.0, -151.928038406)
					(1450.0, -151.812131004)
					(1550.0, -151.690059182)
					(1650.0, -151.635713636)
					(1750.0, -151.528000226)
					(1850.0, -151.452870395)
					(1950.0, -151.372488318)
					(2050.0, -151.317275097)
					(2150.0, -151.24417838)
					(2250.0, -151.195892317)
					(2350.0, -151.138911483)
					(2450.0, -151.074127832)
					(2550.0, -151.020709163)
					(2650.0, -150.979787285)
					(2750.0, -150.930710706)
					(2850.0, -150.895883824)
					(2950.0, -150.818292599)
					(3050.0, -150.788709539)
					(3150.0, -150.734012046)
					(3250.0, -150.70709419)
					(3350.0, -150.677235963)
					(3450.0, -150.651952456)
					(3550.0, -150.611944692)
					(3650.0, -150.59509157)
					(3750.0, -150.559861048)
					(3850.0, -150.551651766)
					(3950.0, -150.516409486)
					(4050.0, -150.504046949)
					(4150.0, -150.460158839)
					(4250.0, -150.455537922)
					(4350.0, -150.445248564)
					(4450.0, -150.4277256)
					(4550.0, -150.416009916)
					(4650.0, -150.418177395)
					(4750.0, -150.395478825)
					(4850.0, -150.368382406)
					(4950.0, -150.345610847)
					(5050.0, -150.3441826)
					(5150.0, -150.330371163)
					(5250.0, -150.318589601)
					(5350.0, -150.31804208)
					(5450.0, -150.293703325)
					(5550.0, -150.273418663)
					(5650.0, -150.25643115)
					(5750.0, -150.244876738)
					(5850.0, -150.238137967)
					(5950.0, -150.211796251)
					(6050.0, -150.204989891)
					(6150.0, -150.183566723)
					(6250.0, -150.174093884)
					(6350.0, -150.162423679)
					(6450.0, -150.160289626)
					(6550.0, -150.138887627)
					(6650.0, -150.133014546)
					(6750.0, -150.120989531)
					(6850.0, -150.122645355)
					(6950.0, -150.11350561)
					(7050.0, -150.111188836)
					(7150.0, -150.090336871)
					(7250.0, -150.090484536)
					(7350.0, -150.087698531)
					(7450.0, -150.070527142)
					(7550.0, -150.07373592)
					(7650.0, -150.077151026)
					(7750.0, -150.066680462)
					(7850.0, -150.054028766)
					(7950.0, -150.038588593)
					(8050.0, -150.039593073)
					(8150.0, -150.035194384)
					(8250.0, -150.03301093)
					(8350.0, -150.034143119)
					(8450.0, -150.015414121)
					(8550.0, -150.00365975)
					(8650.0, -149.998087739)
					(8750.0, -149.995406984)
					(8850.0, -149.996939147)
					(8950.0, -149.983337175)
					(9050.0, -149.977785336)
					(9150.0, -149.969994498)
					(9250.0, -149.968331595)
					(9350.0, -149.959712003)
					(9450.0, -149.962888838)
					(9550.0, -149.952046018)
					(9650.0, -149.951815226)
					(9750.0, -149.949871559)
					(9850.0, -149.947228346)
					(9950.0, -149.941097377)
					(10050.0, -149.939224689)
					(10150.0, -149.926599413)
					(10250.0, -149.92398659)
					(10350.0, -149.927791008)
					(10450.0, -149.917777164)
					(10550.0, -149.916019649)
					(10650.0, -149.9203054)
					(10750.0, -149.916233459)
					(10850.0, -149.910069996)
					(10950.0, -149.903241932)
					(11050.0, -149.902238469)
					(11150.0, -149.898446053)
					(11250.0, -149.90066138)
					(11350.0, -149.902991587)
					(11450.0, -149.889160429)
					(11550.0, -149.881614984)
					(11650.0, -149.87675778)
					(11750.0, -149.874417542)
					(11850.0, -149.877797369)
					(11950.0, -149.868580505)
					(12000.0, -149.859050657)
				};
				\addplot+[
					no marks,
					line width=0pt,
					color={rgb,255:red,0;green,0;blue,0},
					fill={rgb,255:red,0;green,0;blue,0},
					opacity=0.3,
					mark options={solid, fill opacity=0.3},
					forget plot] coordinates {
					(25.0, -164.62622563)
					(50.0, -162.387876726)
					(75.0, -161.438234698)
					(125.0, -160.329817308)
					(175.0, -159.940106193)
					(225.0, -159.851892247)
					(275.0, -159.778118687)
					(325.0, -159.720549523)
					(375.0, -159.687885213)
					(425.0, -159.657290156)
					(475.0, -159.615026649)
					(525.0, -159.610181712)
					(575.0, -159.589042188)
					(625.0, -159.561421678)
					(675.0, -159.558086292)
					(725.0, -159.544968627)
					(775.0, -159.525595357)
					(825.0, -159.536918802)
					(875.0, -159.526966836)
					(925.0, -159.515428833)
					(975.0, -159.517934498)
					(1025.0, -159.5070141)
					(1075.0, -159.494018954)
					(1125.0, -159.499447147)
					(1175.0, -159.494143897)
					(1225.0, -159.486165359)
					(1275.0, -159.490039388)
					(1325.0, -159.481978888)
					(1375.0, -159.475957422)
					(1425.0, -159.485462673)
					(1475.0, -159.485746753)
					(1525.0, -159.483249866)
					(1575.0, -159.487147211)
					(1625.0, -159.480948607)
					(1675.0, -159.477444285)
					(1725.0, -159.482260816)
					(1775.0, -159.477485691)
					(1825.0, -159.473798399)
					(1875.0, -159.474083177)
					(1925.0, -159.466315832)
					(1975.0, -159.46241699)
					(2025.0, -159.467718473)
					(2075.0, -159.4627058)
					(2125.0, -159.459686787)
					(2175.0, -159.459791526)
					(2225.0, -159.451354807)
					(2275.0, -159.44892472)
					(2325.0, -159.454476417)
					(2375.0, -159.449249397)
					(2425.0, -159.446867883)
					(2475.0, -159.44728785)
					(2525.0, -159.439748988)
					(2575.0, -159.437781365)
					(2625.0, -159.442891802)
					(2675.0, -159.437866989)
					(2725.0, -159.436023704)
					(2775.0, -159.436392848)
					(2825.0, -159.429894449)
					(2875.0, -159.428034415)
					(2925.0, -159.432765787)
					(2975.0, -159.428044071)
					(3025.0, -159.426401061)
					(3075.0, -159.426921915)
					(3125.0, -159.420965178)
					(3175.0, -159.419348289)
					(3225.0, -159.424056921)
					(3275.0, -159.419282895)
					(3325.0, -159.417651857)
					(3375.0, -159.418513199)
					(3425.0, -159.412774637)
					(3475.0, -159.411432967)
					(3525.0, -159.415899175)
					(3575.0, -159.411338044)
					(3625.0, -159.409498705)
					(3675.0, -159.410782024)
					(3725.0, -159.405456889)
					(3775.0, -159.403753562)
					(3825.0, -159.408069223)
					(3875.0, -159.403731318)
					(3925.0, -159.401786214)
					(3975.0, -159.403290566)
					(4025.0, -159.398393459)
					(4075.0, -159.396165655)
					(4125.0, -159.400254575)
					(4175.0, -159.395335292)
					(4225.0, -159.39260245)
					(4275.0, -159.393917816)
					(4325.0, -159.389052801)
					(4375.0, -159.38549913)
					(4425.0, -159.388552194)
					(4475.0, -159.38204876)
					(4525.0, -159.378620532)
					(4575.0, -159.378388529)
					(4625.0, -159.373215424)
					(4675.0, -159.369167292)
					(4725.0, -159.369667619)
					(4775.0, -159.36129143)
					(4825.0, -159.355343804)
					(4875.0, -159.351646396)
					(4925.0, -159.345537508)
					(4975.0, -159.336293251)
					(5025.0, -159.330584554)
					(5075.0, -159.322557855)
					(5125.0, -159.31403268)
					(5175.0, -159.303805843)
					(5225.0, -159.291668998)
					(5275.0, -159.278134145)
					(5325.0, -159.274223526)
					(5375.0, -159.265307085)
					(5425.0, -159.256053032)
					(5475.0, -159.24700906)
					(5525.0, -159.236736276)
					(5575.0, -159.229678448)
					(5625.0, -159.229953054)
					(5675.0, -159.224916968)
					(5725.0, -159.22050056)
					(5775.0, -159.214075904)
					(5825.0, -159.208256384)
					(5875.0, -159.204816905)
					(5925.0, -159.207103778)
					(5975.0, -159.204033224)
					(6000.0, -159.976472852)
					(5950.0, -159.976050674)
					(5900.0, -159.975697095)
					(5850.0, -159.976891638)
					(5800.0, -159.978202335)
					(5750.0, -159.973050985)
					(5700.0, -159.973713693)
					(5650.0, -159.972765691)
					(5600.0, -159.971864821)
					(5550.0, -159.971856948)
					(5500.0, -159.972427243)
					(5450.0, -159.965718163)
					(5400.0, -159.966044948)
					(5350.0, -159.963500633)
					(5300.0, -159.960848207)
					(5250.0, -159.959271237)
					(5200.0, -159.958397131)
					(5150.0, -159.95126301)
					(5100.0, -159.951502911)
					(5050.0, -159.949583196)
					(5000.0, -159.94643692)
					(4950.0, -159.946638574)
					(4900.0, -159.947080829)
					(4850.0, -159.941048377)
					(4800.0, -159.943245691)
					(4750.0, -159.941317426)
					(4700.0, -159.939325561)
					(4650.0, -159.941830161)
					(4600.0, -159.943653217)
					(4550.0, -159.938180321)
					(4500.0, -159.940767571)
					(4450.0, -159.93975945)
					(4400.0, -159.93886953)
					(4350.0, -159.941988045)
					(4300.0, -159.943928799)
					(4250.0, -159.938887315)
					(4200.0, -159.94226853)
					(4150.0, -159.941712488)
					(4100.0, -159.941390707)
					(4050.0, -159.944987039)
					(4000.0, -159.947452132)
					(3950.0, -159.942173902)
					(3900.0, -159.945799433)
					(3850.0, -159.945729955)
					(3800.0, -159.94564094)
					(3750.0, -159.949621772)
					(3700.0, -159.952527368)
					(3650.0, -159.946804533)
					(3600.0, -159.950546213)
					(3550.0, -159.950922834)
					(3500.0, -159.950765151)
					(3450.0, -159.955238212)
					(3400.0, -159.958525816)
					(3350.0, -159.952304577)
					(3300.0, -159.956196129)
					(3250.0, -159.957069738)
					(3200.0, -159.956921474)
					(3150.0, -159.96192211)
					(3100.0, -159.965526359)
					(3050.0, -159.958739406)
					(3000.0, -159.962811037)
					(2950.0, -159.964272124)
					(2900.0, -159.964449805)
					(2850.0, -159.969918941)
					(2800.0, -159.973986522)
					(2750.0, -159.96652105)
					(2700.0, -159.970725095)
					(2650.0, -159.972881268)
					(2600.0, -159.973665001)
					(2550.0, -159.97990827)
					(2500.0, -159.984487381)
					(2450.0, -159.976120022)
					(2400.0, -159.980326512)
					(2350.0, -159.983560316)
					(2300.0, -159.985512893)
					(2250.0, -159.992856803)
					(2200.0, -159.99845749)
					(2150.0, -159.98942231)
					(2100.0, -159.993645639)
					(2050.0, -159.999172233)
					(2000.0, -160.002826987)
					(1950.0, -160.011570512)
					(1900.0, -160.021195981)
					(1850.0, -160.012721827)
					(1800.0, -160.020264466)
					(1750.0, -160.028913675)
					(1700.0, -160.036917245)
					(1650.0, -160.053956045)
					(1600.0, -160.078382203)
					(1550.0, -160.076769089)
					(1500.0, -160.097575548)
					(1450.0, -160.118577814)
					(1400.0, -160.157629276)
					(1350.0, -160.181574395)
					(1300.0, -160.206272381)
					(1250.0, -160.196503067)
					(1200.0, -160.21487483)
					(1150.0, -160.223471236)
					(1100.0, -160.246824587)
					(1050.0, -160.263725925)
					(1000.0, -160.275813328)
					(950.0, -160.270353085)
					(900.0, -160.298561423)
					(850.0, -160.311717908)
					(800.0, -160.340702801)
					(750.0, -160.371890534)
					(700.0, -160.397853151)
					(650.0, -160.393553706)
					(600.0, -160.406790334)
					(550.0, -160.421688632)
					(500.0, -160.458301606)
					(450.0, -160.488170643)
					(400.0, -160.529173849)
					(350.0, -160.529938057)
					(300.0, -160.611214313)
					(250.0, -160.810293146)
					(200.0, -161.112743243)
					(150.0, -161.360078442)
					(100.0, -161.611567488)
					(50.0, -163.149691372)
					(25.0, -165.349820996)
				};
				\addplot+[
					no marks,
					line width=1.5pt,
					color={rgb,255:red,0;green,0;blue,0},
					mark options={solid}] coordinates {
					(25.0, -164.988023313)
					(50.0, -162.768784049)
					(75.0, -161.658972461)
					(125.0, -160.889620969)
					(175.0, -160.60124139)
					(225.0, -160.4086722)
					(275.0, -160.238366948)
					(325.0, -160.168685518)
					(375.0, -160.109660304)
					(425.0, -160.073642045)
					(475.0, -160.048707757)
					(525.0, -160.030467616)
					(575.0, -160.000465717)
					(625.0, -159.997934783)
					(675.0, -159.978268614)
					(725.0, -159.956720729)
					(775.0, -159.947586267)
					(825.0, -159.933929992)
					(875.0, -159.911695209)
					(925.0, -159.909664848)
					(975.0, -159.896431909)
					(1025.0, -159.881725385)
					(1075.0, -159.880586279)
					(1125.0, -159.872100104)
					(1175.0, -159.854484564)
					(1225.0, -159.854640218)
					(1275.0, -159.846224384)
					(1325.0, -159.832317682)
					(1375.0, -159.830201855)
					(1425.0, -159.81604619)
					(1475.0, -159.793674765)
					(1525.0, -159.790270068)
					(1575.0, -159.781651265)
					(1625.0, -159.768259317)
					(1675.0, -159.765868425)
					(1725.0, -159.759004218)
					(1775.0, -159.747779905)
					(1825.0, -159.749407365)
					(1875.0, -159.745580573)
					(1925.0, -159.737316152)
					(1975.0, -159.740073102)
					(2025.0, -159.734604716)
					(2075.0, -159.726872865)
					(2125.0, -159.729107889)
					(2175.0, -159.726147499)
					(2225.0, -159.720160457)
					(2275.0, -159.723560107)
					(2325.0, -159.719175661)
					(2375.0, -159.713071196)
					(2425.0, -159.715315401)
					(2475.0, -159.713160203)
					(2525.0, -159.707959651)
					(2575.0, -159.711360192)
					(2625.0, -159.70785581)
					(2675.0, -159.702449347)
					(2725.0, -159.704529949)
					(2775.0, -159.702791778)
					(2825.0, -159.6981264)
					(2875.0, -159.701457808)
					(2925.0, -159.698445862)
					(2975.0, -159.693545169)
					(3025.0, -159.695396443)
					(3075.0, -159.694039559)
					(3125.0, -159.68978444)
					(3175.0, -159.693024991)
					(3225.0, -159.690440824)
					(3275.0, -159.685833575)
					(3325.0, -159.687457575)
					(3375.0, -159.686469223)
					(3425.0, -159.682497912)
					(3475.0, -159.685623411)
					(3525.0, -159.683316645)
					(3575.0, -159.679043292)
					(3625.0, -159.680420453)
					(3675.0, -159.679763294)
					(3725.0, -159.676113039)
					(3775.0, -159.678907346)
					(3825.0, -159.676858368)
					(3875.0, -159.672882289)
					(3925.0, -159.674070931)
					(3975.0, -159.673642097)
					(4025.0, -159.670295088)
					(4075.0, -159.672717708)
					(4125.0, -159.670859509)
					(4175.0, -159.666870283)
					(4225.0, -159.667649418)
					(4275.0, -159.667288888)
					(4325.0, -159.664088722)
					(4375.0, -159.665882829)
					(4425.0, -159.663907729)
					(4475.0, -159.659533837)
					(4525.0, -159.659945767)
					(4575.0, -159.65917654)
					(4625.0, -159.65598605)
					(4675.0, -159.657454013)
					(4725.0, -159.654820388)
					(4775.0, -159.649941944)
					(4825.0, -159.649375312)
					(4875.0, -159.647537158)
					(4925.0, -159.644122762)
					(4975.0, -159.643681927)
					(5025.0, -159.639133551)
					(5075.0, -159.634491778)
					(5125.0, -159.632941496)
					(5175.0, -159.62899744)
					(5225.0, -159.623613396)
					(5275.0, -159.621604289)
					(5325.0, -159.617854772)
					(5375.0, -159.613014621)
					(5425.0, -159.611124193)
					(5475.0, -159.607688935)
					(5525.0, -159.603007688)
					(5575.0, -159.603087214)
					(5625.0, -159.600911859)
					(5675.0, -159.597472986)
					(5725.0, -159.597136586)
					(5775.0, -159.594668083)
					(5825.0, -159.591554895)
					(5875.0, -159.592753203)
					(5925.0, -159.591391574)
					(5975.0, -159.588694674)
					(6000.0, -159.589263692)
				};
			\end{axis}
			\begin{axis}[
				scale only axis,
				width=4cm,
				height=3.5cm,
				at={(5.65cm, 0.0cm)},
				title={Natural image patches},
				xmin=0,
				xmax=20,
				ymin=55,
				ymax=75,
				xlabel={Epochs},
				ylabel={Avg. ELBO $\pm$ 2 $\cdot$ SEM [nat/image]},
				legend entries={MO (B = 500),MO (B = 10k)},
				legend cell align=left,
				legend style={
					at={(0.025,0.975)},
					anchor=north west,
					font=\fontsize{8pt}{8pt}\sansmath\sffamily\selectfont,
					draw=none
				},
				xlabel near ticks,
				ylabel near ticks,
				grid=major,
				xticklabel={\pgfmathprintnumber[precision=4]{\tick}},
				yticklabel={\pgfmathprintnumber[precision=4]{\tick}},
				zticklabel={\pgfmathprintnumber[precision=4]{\tick}},
				title style={font=\fontsize{8pt}{8pt}\sansmath\sffamily\selectfont},
				label style={font=\fontsize{8pt}{8pt}\sansmath\sffamily\selectfont},
				tick label style={font=\fontsize{8pt}{8pt}\sansmath\sffamily\selectfont},
				axis x line=bottom,
				axis y line=left]
				\addplot+[
					no marks,
					line width=0pt,
					color={rgb,255:red,180;green,140;blue,10},
					fill={rgb,255:red,180;green,140;blue,10},
					opacity=0.3,
					mark options={solid, fill opacity=0.3},
					forget plot] coordinates {
					(0.002, -573.644120091)
					(0.003, 34.4318618504)
					(1.0, 61.2224382497)
					(2.0, 61.7437303034)
					(4.0, 62.2836742589)
					(5.0, 62.3658222453)
					(8.0, 62.4172435793)
					(10.0, 62.648111161)
					(15.0, 62.9114975078)
					(20.0, 62.9317903832)
					(20.0, 59.2028306382)
					(15.0, 59.1999484495)
					(10.0, 59.2029852584)
					(8.0, 59.186910231)
					(5.0, 59.1416924637)
					(4.0, 59.1013648679)
					(2.0, 58.8757182846)
					(1.0, 58.6230222558)
					(0.003, 8.98475225125)
					(0.002, -759.087900818)
				};
				\addplot+[
					no marks,
					line width=1.5pt,
					color={rgb,255:red,180;green,140;blue,10},
					mark options={solid}] coordinates {
					(0.002, -666.366010455)
					(0.003, 21.7083070508)
					(1.0, 59.9227302527)
					(2.0, 60.309724294)
					(4.0, 60.6925195634)
					(5.0, 60.7537573545)
					(8.0, 60.8020769051)
					(10.0, 60.9255482097)
					(15.0, 61.0557229786)
					(20.0, 61.0673105107)
				};
				\addplot+[
					no marks,
					line width=0pt,
					color={rgb,255:red,180;green,100;blue,40},
					fill={rgb,255:red,180;green,100;blue,40},
					opacity=0.3,
					mark options={solid, fill opacity=0.3},
					forget plot] coordinates {
					(0.04, 55.0619149731)
					(0.06, 59.0370711297)
					(1.0, 64.8494004776)
					(2.0, 65.9117900548)
					(4.0, 66.5974489925)
					(5.0, 66.7313818004)
					(8.0, 66.9640515464)
					(10.0, 67.0603765372)
					(15.0, 67.1507581436)
					(20.0, 67.1977503113)
					(20.0, 67.0582539937)
					(15.0, 67.0194535694)
					(10.0, 66.913477016)
					(8.0, 66.8480167919)
					(5.0, 66.6454117558)
					(4.0, 66.5227197793)
					(2.0, 65.8922885257)
					(1.0, 64.7807180025)
					(0.06, 58.8233681456)
					(0.04, 54.7612280096)
				};
				\addplot+[
					no marks,
					line width=1.5pt,
					color={rgb,255:red,180;green,100;blue,40},
					mark options={solid}] coordinates {
					(0.04, 54.9115714913)
					(0.06, 58.9302196376)
					(1.0, 64.81505924)
					(2.0, 65.9020392902)
					(4.0, 66.5600843859)
					(5.0, 66.6883967781)
					(8.0, 66.9060341691)
					(10.0, 66.9869267766)
					(15.0, 67.0851058565)
					(20.0, 67.1280021525)
				};
				\addplot+[
					no marks,
					line width=0pt,
					color={rgb,255:red,45;green,158;blue,250},
					fill={rgb,255:red,45;green,158;blue,250},
					opacity=0.3,
					mark options={solid, fill opacity=0.3},
					forget plot] coordinates {
					(0.0, -1773.24468009)
					(0.2, 62.412657441)
					(0.4, 63.7332940764)
					(0.8, 64.5903364081)
					(1.2, 65.080304662)
					(1.6, 65.3626711066)
					(2.0, 65.5665278899)
					(2.4, 65.7287067332)
					(2.8, 65.8680436053)
					(3.2, 65.9689514744)
					(3.6, 66.0413469272)
					(4.0, 66.1113821552)
					(4.4, 66.1817992028)
					(4.8, 66.2384332007)
					(5.2, 66.3156308694)
					(5.6, 66.3749425496)
					(6.0, 66.4346131474)
					(6.4, 66.4889820991)
					(6.8, 66.5014197939)
					(7.2, 66.5547284752)
					(7.6, 66.5685296333)
					(8.0, 66.6118398302)
					(8.4, 66.6364929243)
					(8.8, 66.6432602207)
					(9.2, 66.6753834752)
					(9.6, 66.6780427171)
					(10.0, 66.7083065442)
					(9.8, 66.5036888549)
					(9.4, 66.4877667005)
					(9.0, 66.462219328)
					(8.6, 66.4448689812)
					(8.2, 66.4303609682)
					(7.8, 66.4145652217)
					(7.4, 66.3887612243)
					(7.0, 66.3523066374)
					(6.6, 66.3129416955)
					(6.2, 66.3071369148)
					(5.8, 66.2612278922)
					(5.4, 66.2127161426)
					(5.0, 66.1718526441)
					(4.6, 66.1016349302)
					(4.2, 66.0634399417)
					(3.8, 65.9897807333)
					(3.4, 65.9363395611)
					(3.0, 65.8538092486)
					(2.6, 65.7223805657)
					(2.2, 65.5892071653)
					(1.8, 65.3738000287)
					(1.4, 65.1382463097)
					(1.0, 64.8317282519)
					(0.8, 64.5666364937)
					(0.6, 64.187488114)
					(0.4, 63.5378099383)
					(0.2, 62.1592699547)
					(0.0, -1773.29873573)
				};
				\addplot+[
					no marks,
					line width=1.5pt,
					color={rgb,255:red,45;green,158;blue,250},
					mark options={solid},
					forget plot] coordinates {
					(0.0, -1773.27170791)
					(0.2, 62.2859636978)
					(0.4, 63.6355520073)
					(0.6, 64.2674514086)
					(0.8, 64.5784864509)
					(1.0, 64.8595983467)
					(1.4, 65.1887736185)
					(1.8, 65.4173011351)
					(2.2, 65.6285966785)
					(2.6, 65.7675693918)
					(3.0, 65.8793033396)
					(3.4, 65.9654080564)
					(3.8, 66.0317087719)
					(4.2, 66.1053406079)
					(4.6, 66.1556633826)
					(5.0, 66.2290576635)
					(5.4, 66.2855222874)
					(5.8, 66.3310210468)
					(6.2, 66.3886569256)
					(6.6, 66.4008638732)
					(7.0, 66.4411716137)
					(7.4, 66.4784976917)
					(7.8, 66.4992397357)
					(8.2, 66.5282501742)
					(8.6, 66.5388092009)
					(9.0, 66.5654997941)
					(9.4, 66.5852465713)
					(9.8, 66.5945459983)
					(10.0, 66.6057752491)
				};
				\addplot+[
					no marks,
					line width=0pt,
					color={rgb,255:red,0;green,0;blue,0},
					fill={rgb,255:red,0;green,0;blue,0},
					opacity=0.3,
					mark options={solid, fill opacity=0.3},
					forget plot] coordinates {
					(0.0, -1772.71128283)
					(0.4, 58.3841449515)
					(0.8, 59.9133517735)
					(1.2, 60.0644891024)
					(1.6, 60.1005434686)
					(2.0, 60.0274196454)
					(2.4, 60.1339277802)
					(2.8, 60.142786351)
					(3.2, 60.1535367872)
					(3.6, 60.1581335623)
					(4.0, 60.09023511)
					(4.4, 60.1672851517)
					(4.8, 60.1697428147)
					(5.2, 60.1752385894)
					(5.6, 60.1768589686)
					(6.0, 60.1177743672)
					(6.4, 60.1819313436)
					(6.8, 60.1827348909)
					(7.2, 60.1860484193)
					(7.6, 60.1872129647)
					(8.0, 60.1346678523)
					(8.4, 60.1906603608)
					(8.8, 60.1910159996)
					(9.2, 60.1927788533)
					(9.6, 60.1940280628)
					(10.0, 60.1464953185)
					(10.4, 60.1965827985)
					(10.8, 60.1966880468)
					(11.2, 60.197461368)
					(11.6, 60.1990726458)
					(12.0, 60.1553795725)
					(12.4, 60.20094084)
					(12.8, 60.2008865316)
					(13.2, 60.200962146)
					(13.6, 60.2030313906)
					(14.0, 60.1624085104)
					(14.4, 60.2043162233)
					(14.8, 60.204192775)
					(15.2, 60.2037157608)
					(15.6, 60.2062762354)
					(16.0, 60.1681050866)
					(16.4, 60.2070171373)
					(16.8, 60.2069292216)
					(17.2, 60.2059592235)
					(17.6, 60.2089617784)
					(18.0, 60.1728241489)
					(18.4, 60.209228415)
					(18.8, 60.2092132525)
					(19.2, 60.2078309098)
					(19.6, 60.2111993824)
					(20.0, 60.1768195589)
					(20.0, 58.3494721318)
					(19.6, 58.392934158)
					(19.2, 58.3862866513)
					(18.8, 58.394167046)
					(18.4, 58.3902371327)
					(18.0, 58.3449574875)
					(17.6, 58.3907161088)
					(17.2, 58.3845481195)
					(16.8, 58.3919598823)
					(16.4, 58.3882672476)
					(16.0, 58.3396009424)
					(15.6, 58.3877902771)
					(15.2, 58.3824808434)
					(14.8, 58.3893305342)
					(14.4, 58.3858652971)
					(14.0, 58.3330840934)
					(13.6, 58.3843024896)
					(13.2, 58.3799744336)
					(12.8, 58.386296901)
					(12.4, 58.3828812852)
					(12.0, 58.3250193832)
					(11.6, 58.3803106843)
					(11.2, 58.3768248802)
					(10.8, 58.3824755978)
					(10.4, 58.3790387839)
					(10.0, 58.3149466325)
					(9.6, 58.3754132449)
					(9.2, 58.3726376485)
					(8.8, 58.3771644618)
					(8.4, 58.3737948785)
					(8.0, 58.3015946431)
					(7.6, 58.369062525)
					(7.2, 58.3666314235)
					(6.8, 58.3690559159)
					(6.4, 58.3659458635)
					(6.0, 58.2827265488)
					(5.6, 58.3595434)
					(5.2, 58.356967397)
					(4.8, 58.357058201)
					(4.4, 58.3527194039)
					(4.0, 58.2522447464)
					(3.6, 58.3422823003)
					(3.2, 58.3372867414)
					(2.8, 58.3308450933)
					(2.4, 58.3211847964)
					(2.0, 58.182956558)
					(1.6, 58.2828223093)
					(1.2, 58.2509325103)
					(0.8, 58.1560182295)
					(0.4, 58.3547523538)
					(0.0, -1773.37725996)
				};
				\addplot+[
					no marks,
					line width=1.5pt,
					color={rgb,255:red,0;green,0;blue,0},
					mark options={solid},
					forget plot] coordinates {
					(0.0, -1773.04427139)
					(0.4, 58.3694486526)
					(0.8, 59.0346850015)
					(1.2, 59.1577108064)
					(1.6, 59.1916828889)
					(2.0, 59.1051881017)
					(2.4, 59.2275562883)
					(2.8, 59.2368157222)
					(3.2, 59.2454117643)
					(3.6, 59.2502079313)
					(4.0, 59.1712399282)
					(4.4, 59.2600022778)
					(4.8, 59.2634005078)
					(5.2, 59.2661029932)
					(5.6, 59.2682011843)
					(6.0, 59.200250458)
					(6.4, 59.2739386036)
					(6.8, 59.2758954034)
					(7.2, 59.2763399214)
					(7.6, 59.2781377449)
					(8.0, 59.2181312477)
					(8.4, 59.2822276196)
					(8.8, 59.2840902307)
					(9.2, 59.2827082509)
					(9.6, 59.2847206538)
					(10.0, 59.2307209755)
					(10.4, 59.2878107912)
					(10.8, 59.2895818223)
					(11.2, 59.2871431241)
					(11.6, 59.289691665)
					(12.0, 59.2401994778)
					(12.4, 59.2919110626)
					(12.8, 59.2935917163)
					(13.2, 59.2904682898)
					(13.6, 59.2936669401)
					(14.0, 59.2477463019)
					(14.4, 59.2950907602)
					(14.8, 59.2967616546)
					(15.2, 59.2930983021)
					(15.6, 59.2970332562)
					(16.0, 59.2538530145)
					(16.4, 59.2976421925)
					(16.8, 59.2994445519)
					(17.2, 59.2952536715)
					(17.6, 59.2998389436)
					(18.0, 59.2588908182)
					(18.4, 59.2997327738)
					(18.8, 59.3016901493)
					(19.2, 59.2970587806)
					(19.6, 59.3020667702)
					(20.0, 59.2631458454)
				};
			\end{axis}
		\end{tikzpicture}
			\caption{Mixture modeling results. \textbf{A}. Expected parameters, $\mathbb{E}[\bm{\beta}]$, under
			the mean-field solutions found via trust-region updates (top) and natural gradient steps (bottom).
			\textbf{B}. Lower bound averaged over multiple runs as a function of the number of updates
			of the parameters. In case of the trust-region method, each step of the inner loop is counted
			as one update. \textbf{C}. Lower bound for $8 \times 8$ natural image patches sampled from the van
			Hateren dataset. The horizontal axis indicates the number of passes through the dataset.
			An update of the trust-region method takes twice as long as a natural gradient update.
			A comparison is made to memoized variational inference (MO) with batch sizes of 500 and
			10,000. Error bars of TR and MO (B = 10k) are too small to be visible.}
			\label{fig:mixtures}
		\end{figure*}
		As a first illustrative example, consider a mixture of multivariate Bernoulli distributions with $K$ components,
		where the global parameters are given by component parameters $\bm{\beta}_k \in \left[ 0, 1 \right]^D$ and mixture weights $\bm{\pi}$.
		We assume uniform prior distributions on $\bm{\beta}_k$ and $\bm{\pi}$ and beta and Dirichlet distributions
		for the approximate posterior distributions,
		\begin{align*}
			\textstyle
			q(\bm{\beta}) \propto \prod_{k,i} \beta_{ki}^{a_{ki} - 1}(1 - \beta_{ki})^{b_{ki} - 1}, \quad
			q(\bm{\pi}) \propto \prod_k \pi_k^{\gamma_k - 1}.
		\end{align*}
		Using 40 components, we applied this model to a binarized version of MNIST where each pixel has been randomly
		set to 0 or 1 according to its grayscale value interpreted as a probability. The $a_{ki}$ and $b_{ki}$ were randomly
		initialized by sampling from a gamma distribution with shape parameter
		100 and scale parameter 0.01, and the $\gamma_k$ were initialized at 1. We ran SVI with
		trust-region updates for 10 epochs (passes through the dataset) using 2 iterations for the inner loop (Algorithm~\ref{alg:trsvi}) and SVI with
		natural gradient steps for 20 epochs. For both methods we used a batch size of 200 and the learning rate schedule
		$\rho_t = (\tau + t)^{-\kappa}$,
		where for $\kappa$ we used 0.5 and for $\tau$ we used 100.
		
		Figure~\ref{fig:mixtures}A shows the expected values of the probabilities $\beta_{ki}$ under the posterior approximations
		found by the two methods. The solution found via natural gradient steps makes use of less than half of the mixture components,
		leading to significantly poorer performance (Figure~\ref{fig:mixtures}B).
		This can be explained as follows. A mixture component which is inconsistent with the data
		will be assigned only a few data points.
		After updating $\gamma$ and $\bm{\lambda}_k$, this mixture component will have lower prior
		probability and will shrink towards the prior distribution, which
		makes it less consistent with the data and in turn leads to the assignment of even fewer data points.
		The trust-region updates alleviate this problem by initializing $\phi_{nk}$ with $1/K$, that is,
		assigning the data points in the current batch equally to all mixture components. This forces the mixture components
		to initially become more consistent with the current batch of data.

	\subsection{Natural image patches}
		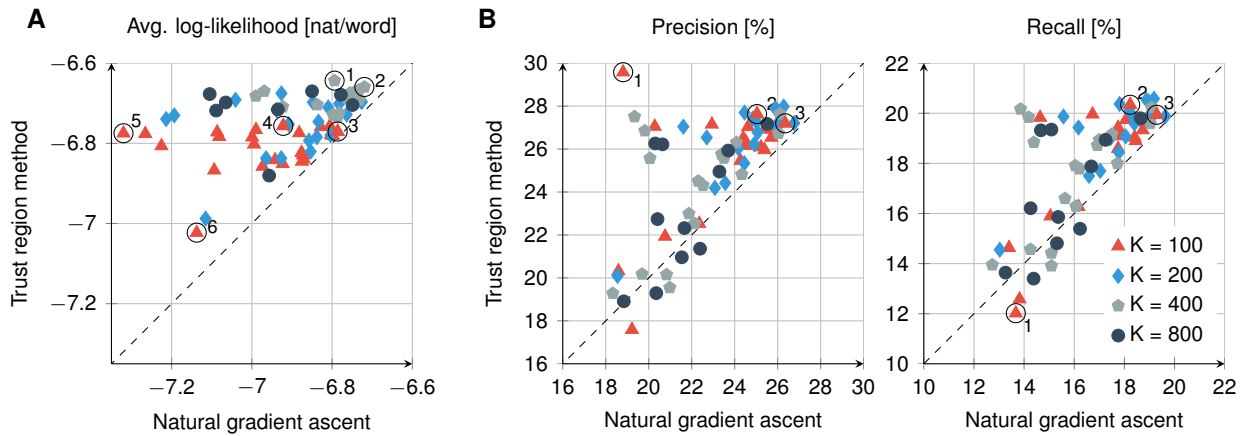
\begin{figure*}[t]
			\centering
			\begin{tikzpicture}
				\node at (-7cm, 4.6cm) {\sffamily\textbf A};
				\node at (-1cm, 4.6cm) {\sffamily\textbf B};

				\begin{axis}[
					scale only axis,
					width=4cm,
					height=4cm,
					at={(-6.0cm, 0.0cm)},
					title={Avg. log-likelihood [nat/word]},
					xmin=-7.35,
					xmax=-6.6,
					ymin=-7.35,
					ymax=-6.6,
					xlabel={Natural gradient ascent},
					ylabel={Trust region method},
					xlabel near ticks,
					ylabel near ticks,
					grid=major,
					xtick={-7.2,-7.0,-6.8,-6.6},
					ytick={-7.2,-7.0,-6.8,-6.6},
					xticklabel={\pgfmathprintnumber[precision=4]{\tick}},
					yticklabel={\pgfmathprintnumber[precision=4]{\tick}},
					zticklabel={\pgfmathprintnumber[precision=4]{\tick}},
					title style={font=\fontsize{8pt}{8pt}\sansmath\sffamily\selectfont},
					label style={font=\fontsize{8pt}{8pt}\sansmath\sffamily\selectfont},
					tick label style={font=\fontsize{8pt}{8pt}\sansmath\sffamily\selectfont},
					axis x line=bottom,
					axis y line=left]
					\addplot+[no marks, dashed, black, mark options={solid}, forget plot] coordinates {
						(-8.0, -8.0)
						(-6.0, -6.0)
					};
					\addplot+[
						only marks,
						color={rgb,1:red,0.91;green,0.30;blue,0.24},
						mark=triangle*,
						mark options={solid, scale=1.4}] coordinates {
						(-6.797449, -6.781205)
						(-6.805334, -6.756594)
						(-6.803275, -6.779005)
						(-6.921754, -6.756444)
						(-6.828114, -6.759154)
						(-7.093826, -6.867089)
						(-6.876222, -6.823182)
						(-7.32, -6.774841)
						(-6.942086, -6.84146)
						(-7.086437, -6.77205)
						(-6.923267, -6.850885)
						(-7.265642, -6.776158)
						(-6.882015, -6.774192)
						(-7.081455, -6.783428)
						(-6.994487, -6.802449)
						(-6.78612, -6.770606)
						(-6.973148, -6.857849)
						(-7.137453, -7.022782)
						(-6.875074, -6.845206)
						(-6.872654, -6.829445)
						(-6.870385, -6.839747)
						(-7.22524, -6.80671)
						(-6.989642, -6.766254)
						(-6.999167, -6.784835)
					};
					\addplot+[
						only marks,
						color={rgb,1:red,0.20;green,0.60;blue,0.86},
						mark=diamond*,
						mark options={solid, scale=1.4}] coordinates {
						(-7.115299, -6.987694)
						(-6.725746, -6.696634)
						(-6.926259, -6.675299)
						(-6.809546, -6.71042)
						(-6.728496, -6.694257)
						(-6.926718, -6.837168)
						(-6.833915, -6.745894)
						(-6.847962, -6.697409)
						(-6.776558, -6.684105)
						(-7.193386, -6.730228)
						(-6.853328, -6.820572)
						(-7.213603, -6.739482)
						(-6.857461, -6.793267)
						(-6.763636, -6.728141)
						(-6.907915, -6.752363)
						(-6.964582, -6.837221)
						(-7.040555, -6.691443)
						(-6.803695, -6.779359)
						(-6.938424, -6.716866)
						(-6.782401, -6.701235)
						(-6.837918, -6.784411)
					};
					\addplot+[
						only marks,
						color={rgb,1:red,0.58;green,0.65;blue,0.65},
						mark=pentagon*,
						mark options={solid, scale=1.2}] coordinates {
						(-6.732358, -6.691438)
						(-6.752001, -6.688782)
						(-6.755147, -6.674179)
						(-6.76323, -6.686557)
						(-6.786262, -6.734472)
						(-6.990031, -6.682046)
						(-6.797506, -6.728281)
						(-6.781954, -6.724726)
						(-6.836808, -6.703765)
						(-6.793717, -6.643597)
						(-6.923115, -6.708866)
						(-6.760215, -6.687535)
						(-6.744875, -6.672395)
						(-6.733256, -6.666802)
						(-6.969103, -6.669936)
						(-6.719524, -6.659395)
					};
					\addplot+[
						only marks,
						color={rgb,1:red,0.20;green,0.29;blue,0.37},
						mark=*,
						mark options={solid, scale=1.2}] coordinates {
						(-6.748944, -6.704337)
						(-6.849939, -6.670126)
						(-7.088986, -6.71824)
						(-6.777987, -6.678757)
						(-7.10508, -6.677027)
						(-7.065163, -6.698078)
						(-6.93495, -6.715529)
						(-6.956858, -6.88047)
					};
					\addplot+[
						only marks,
						color={rgb,1:red,0.91;green,0.30;blue,0.24},
						mark=triangle*,
						mark options={solid, scale=1.4},
						forget plot] coordinates {
						(-6.921754, -6.756444)
						(-6.78612, -6.770606)
					};
					\addplot+[only marks, black, mark=o, mark options={solid, scale=1.8}] coordinates {
						(-6.793717, -6.643597)
					};
					\node at (axis cs:-6.753717,-6.627597) {\scriptsize\sffamily 1};
					\addplot+[only marks, black, mark=o, mark options={solid, scale=1.8}] coordinates {
						(-6.719524, -6.659395)
					};
					\node at (axis cs:-6.679524,-6.643395) {\scriptsize\sffamily 2};
					\addplot+[only marks, black, mark=o, mark options={solid, scale=1.8}] coordinates {
						(-6.78612, -6.770606)
					};
					\node at (axis cs:-6.74612,-6.754606) {\scriptsize\sffamily 3};
					\addplot+[only marks, black, mark=o, mark options={solid, scale=1.8}] coordinates {
						(-6.921754, -6.756444)
					};
					\node at (axis cs:-6.960754,-6.748444) {\scriptsize\sffamily 4};
					\addplot+[only marks, black, mark=o, mark options={solid, scale=1.8}] coordinates {
						(-7.32, -6.774841)
					};
					\node at (axis cs:-7.288,-6.743841) {\scriptsize\sffamily 5};
					\addplot+[only marks, black, mark=o, mark options={solid, scale=1.8}] coordinates {
						(-7.137453, -7.022782)
					};
					\node at (axis cs:-7.097453,-7.006782) {\scriptsize\sffamily 6};
				\end{axis}

				\begin{axis}[
					scale only axis,
					width=4cm,
					height=4cm,
					at={(4.8cm, 0.0cm)},
					title={Recall [\%]},
					xmin=10,
					xmax=22,
					ymin=10,
					ymax=22,
					xlabel={Natural gradient ascent},
					legend entries={K = 100,K = 200,K = 400,K = 800},
					legend cell align=left,
					legend style={
						at={(0.975,0.025)},
						anchor=south east,
						font=\fontsize{8pt}{8pt}\sansmath\sffamily\selectfont,
						draw=none
					},
					xlabel near ticks,
					ylabel near ticks,
					grid=major,
					xtick={10,12,14,16,18,20,22},
					ytick={10,12,14,16,18,20,22},
					xticklabel={\pgfmathprintnumber[precision=4]{\tick}},
					yticklabel={\pgfmathprintnumber[precision=4]{\tick}},
					zticklabel={\pgfmathprintnumber[precision=4]{\tick}},
					title style={font=\fontsize{8pt}{8pt}\sansmath\sffamily\selectfont},
					label style={font=\fontsize{8pt}{8pt}\sansmath\sffamily\selectfont},
					tick label style={font=\fontsize{8pt}{8pt}\sansmath\sffamily\selectfont},
					axis x line=bottom,
					axis y line=left]
					\addplot+[no marks, dashed, black, mark options={solid}, forget plot] coordinates {
						(10.0, 10.0)
						(40.0, 40.0)
					};
					\addplot+[
						only marks,
						color={rgb,1:red,0.91;green,0.30;blue,0.24},
						mark=triangle*,
						mark options={solid, scale=1.4}] coordinates {
						(18.6, 19.76)
						(14.64, 19.83)
						(13.67, 12.02)
						(13.4, 14.63)
						(18.41, 19.04)
						(18.72, 19.65)
						(18.23, 20.34)
						(17.75, 19.43)
						(17.82, 19.28)
						(19.29, 19.94)
						(15.05, 15.89)
						(18.06, 19.74)
						(16.17, 16.27)
						(17.75, 18.59)
						(13.82, 12.58)
						(18.73, 19.33)
						(16.73, 19.96)
						(18.45, 18.9)
						(17.79, 19.09)
					};
					\addplot+[
						only marks,
						color={rgb,1:red,0.20;green,0.60;blue,0.86},
						mark=diamond*,
						mark options={solid, scale=1.4}] coordinates {
						(18.98, 20.46)
						(17.78, 18.44)
						(18.29, 19.96)
						(18.92, 19.81)
						(17.45, 19.01)
						(16.59, 17.5)
						(18.88, 20.54)
						(12.59, 9.87)
						(16.17, 19.43)
						(18.19, 20.09)
						(19.18, 20.58)
						(13.03, 14.55)
						(18.4, 19.59)
						(17.81, 20.37)
						(19.64, 19.88)
						(17.04, 17.7)
						(15.57, 19.87)
						(18.05, 19.08)
						(19.13, 19.9)
					};
					\addplot+[
						only marks,
						color={rgb,1:red,0.58;green,0.65;blue,0.65},
						mark=pentagon*,
						mark options={solid, scale=1.2}] coordinates {
						(12.73, 13.95)
						(16.07, 16.29)
						(16.92, 18.72)
						(16.04, 17.91)
						(14.25, 14.57)
						(14.18, 19.84)
						(14.39, 18.84)
						(15.63, 16.6)
						(16.21, 17.78)
						(17.72, 18.0)
						(17.0, 18.94)
						(19.03, 19.59)
						(15.09, 13.91)
						(19.08, 20.23)
						(17.51, 19.17)
						(13.86, 20.17)
						(15.09, 14.42)
					};
					\addplot+[
						only marks,
						color={rgb,1:red,0.20;green,0.29;blue,0.37},
						mark=*,
						mark options={solid, scale=1.2}] coordinates {
						(13.26, 13.64)
						(16.68, 17.88)
						(14.26, 16.21)
						(18.65, 19.8)
						(15.09, 19.35)
						(14.38, 13.4)
						(16.23, 15.39)
						(17.25, 18.94)
						(15.36, 15.86)
						(15.31, 14.81)
						(14.67, 19.32)
					};
					\addplot+[
						only marks,
						color={rgb,1:red,0.91;green,0.30;blue,0.24},
						mark=triangle*,
						mark options={solid, scale=1.4},
						forget plot] coordinates {
						(18.23, 20.34)
					};
					\addplot+[
						only marks,
						color={rgb,1:red,0.91;green,0.30;blue,0.24},
						mark=triangle*,
						mark options={solid, scale=1.4},
						forget plot] coordinates {
						(19.29, 19.94)
					};
					\addplot+[only marks, black, mark=o, mark options={solid, scale=1.8}] coordinates {
						(13.67, 12.02)
					};
					\node at (axis cs:14.17,11.52) {\scriptsize\sffamily 1};
					\addplot+[only marks, black, mark=o, mark options={solid, scale=1.8}] coordinates {
						(18.23, 20.34)
					};
					\node at (axis cs:18.73,20.84) {\scriptsize\sffamily 2};
					\addplot+[only marks, black, mark=o, mark options={solid, scale=1.8}] coordinates {
						(19.29, 19.94)
					};
					\node at (axis cs:19.79,20.44) {\scriptsize\sffamily 3};
				\end{axis}
				\begin{axis}[
					scale only axis,
					width=4cm,
					height=4cm,
					at={(0.0cm, 0.0cm)},
					title={Precision [\%]},
					xmin=16,
					xmax=30,
					ymin=16,
					ymax=30,
					xlabel={Natural gradient ascent},
					ylabel={Trust region method},
					xlabel near ticks,
					ylabel near ticks,
					grid=major,
					xtick={16,18,20,22,24,26,28,30},
					ytick={16,18,20,22,24,26,28,30},
					xticklabel={\pgfmathprintnumber[precision=4]{\tick}},
					yticklabel={\pgfmathprintnumber[precision=4]{\tick}},
					zticklabel={\pgfmathprintnumber[precision=4]{\tick}},
					title style={font=\fontsize{8pt}{8pt}\sansmath\sffamily\selectfont},
					label style={font=\fontsize{8pt}{8pt}\sansmath\sffamily\selectfont},
					tick label style={font=\fontsize{8pt}{8pt}\sansmath\sffamily\selectfont},
					axis x line=bottom,
					axis y line=left]
					\addplot+[no marks, dashed, black, mark options={solid}, forget plot] coordinates {
						(10.0, 10.0)
						(40.0, 40.0)
					};
					\addplot+[
						only marks,
						color={rgb,1:red,0.91;green,0.30;blue,0.24},
						mark=triangle*,
						mark options={solid, scale=1.4}] coordinates {
						(25.42, 26.98)
						(20.28, 27.05)
						(18.81, 29.58)
						(18.59, 20.32)
						(25.22, 26.09)
						(25.68, 26.91)
						(25.04, 27.63)
						(24.4, 26.51)
						(24.51, 26.35)
						(26.35, 27.2)
						(20.76, 21.93)
						(24.62, 27.03)
						(22.35, 22.53)
						(24.27, 25.46)
						(19.22, 17.59)
						(25.71, 26.54)
						(22.94, 27.15)
						(25.39, 25.96)
						(24.49, 26.16)
					};
					\addplot+[
						only marks,
						color={rgb,1:red,0.20;green,0.60;blue,0.86},
						mark=diamond*,
						mark options={solid, scale=1.4}] coordinates {
						(25.99, 27.8)
						(24.45, 25.34)
						(25.01, 27.27)
						(25.91, 27.08)
						(24.17, 26.04)
						(23.08, 24.19)
						(25.89, 27.89)
						(17.47, 14.56)
						(22.7, 26.54)
						(24.85, 27.42)
						(26.28, 27.98)
						(18.55, 20.09)
						(25.07, 26.78)
						(24.46, 27.7)
						(26.77, 27.21)
						(23.56, 24.42)
						(21.61, 27.04)
						(24.91, 26.23)
						(26.11, 27.21)
					};
					\addplot+[
						only marks,
						color={rgb,1:red,0.58;green,0.65;blue,0.65},
						mark=pentagon*,
						mark options={solid, scale=1.2}] coordinates {
						(18.33, 19.28)
						(22.14, 22.53)
						(23.5, 25.58)
						(22.3, 24.51)
						(19.69, 20.18)
						(19.82, 26.85)
						(20.06, 25.57)
						(21.87, 22.99)
						(22.53, 24.31)
						(24.33, 24.82)
						(23.39, 25.76)
						(26.04, 26.73)
						(20.98, 19.55)
						(26.1, 27.64)
						(24.08, 26.3)
						(19.34, 27.51)
						(20.83, 20.15)
					};
					\addplot+[
						only marks,
						color={rgb,1:red,0.20;green,0.29;blue,0.37},
						mark=*,
						mark options={solid, scale=1.2}] coordinates {
						(18.84, 18.91)
						(23.29, 24.95)
						(20.41, 22.74)
						(25.51, 27.17)
						(20.64, 26.22)
						(20.35, 19.29)
						(22.39, 21.36)
						(23.7, 25.93)
						(21.66, 22.32)
						(21.54, 20.96)
						(20.3, 26.26)
					};
					\addplot+[
						only marks,
						color={rgb,1:red,0.91;green,0.30;blue,0.24},
						mark=triangle*,
						mark options={solid, scale=1.4},
						forget plot] coordinates {
						(25.04, 27.63)
					};
					\addplot+[
						only marks,
						color={rgb,1:red,0.91;green,0.30;blue,0.24},
						mark=triangle*,
						mark options={solid, scale=1.4},
						forget plot] coordinates {
						(26.35, 27.2)
					};
					\addplot+[only marks, black, mark=o, mark options={solid, scale=1.8}] coordinates {
						(18.81, 29.58)
					};
					\node at (axis cs:19.51,29.18) {\scriptsize\sffamily 1};
					\addplot+[only marks, black, mark=o, mark options={solid, scale=1.8}] coordinates {
						(25.04, 27.63)
					};
					\node at (axis cs:25.74,28.13) {\scriptsize\sffamily 2};
					\addplot+[only marks, black, mark=o, mark options={solid, scale=1.8}] coordinates {
						(26.35, 27.2)
					};
					\node at (axis cs:27.05,27.7) {\scriptsize\sffamily 3};
				\end{axis}
			\end{tikzpicture}
			\caption{Hyperparameter search results. The circled results' hyperparameters are listed
				in Table~\ref{tbl:wikipedia} and Table~\ref{tbl:netflix}.
				\textbf{A}. Performance on a validation set of Wikipedia articles.
				Shown are estimated log-likelihoods of the expected parameters found via
				the trust-region method and natural gradient ascent using randomly chosen hyperparameters.
				\textbf{B}. Performance on a validation set of Netflix users.}
			\label{fig:parameter_search}
		\end{figure*}
		We next considered the task of modeling natural image patches sampled from the van Hateren
		image dataset \cite{vanHateren:1998}. We used $8 \times 8$ image patches, which is a
		commonly used size in applications such as image compression or restoration
		\citep[e.g.,][]{Zoran:2011}.
		We fit mixtures of Gaussians with 50 components and normal-inverse-Wishart priors on the
		means and covariances of the components to 500,000 image patches. We used a batch size of
		$500$, $\kappa = .5$, $\tau = 10$, and a fixed number of 5 iterations for the inner loop of
		the trust-region method. For comparison we also trained models with \textit{memoized
		variational inference} \citep[MO; ][]{Hughes:2013} using the same settings for the priors and
		the same initialization of the distributions as with our methods. MO is an online
		inference algorithm akin to gradient averaging. Its updates are less noisy than those of
		SVI but it requires memorizing each individual update, generating a considerable memory overhead
		when the dimensionality of the gradients or the number of batches is large.

		Figure~\ref{fig:mixtures} shows that SVI with natural gradient steps quickly gets stuck. MO
		with the same batch size performs slightly better, and MO with a large batch size of
		10,000 does very well. Our trust-region method with a batch size of 500 achieves comparable performance
		and uses much less memory.

	\subsection{Wikipedia articles}
		We applied our algorithm to latent Dirichlet allocation \citep[LDA;][]{Blei:2003} and 
		a dataset of approximately 3.8M Wikipedia articles. The model is
		\begin{align}
			\textstyle
			p(\bm{\beta}) &\propto \prod_{ki} \beta_{ki}^{\eta - 1}, &
			p(\bm{\theta}_n) &\propto \prod_k \theta_{nk}^{\alpha_k - 1},
		\end{align}
		\vspace{-.6cm}
		\begin{align}
			p(x_{nm}, z_{nm} \mid \bm{\theta}_n, \bm{\beta}) &= \theta_{nz_{nm}}\beta_{z_{nm}x_{nm}}.
		\end{align}
		Figure~\ref{fig:parameter_search}A shows the performance of our trust-region method plotted against natural gradient
		ascent for randomly selected hyperparameters. The parameters were selected from a grid with
		the batch size $B$ and learning rate parameters selected from
		\begin{align*}
			\textstyle
			B \in \{10, 50, 100, 500, 1000\}, \\
			\kappa \in \{ 0.5, 0.6, 0.7, 0.8, 0.9, 1.0 \}, \\
			\tau \in \{ 1, 10, 100, 1000, 10000 \}.
		\end{align*}
		For the trust-region method we used $m$ steps in the inner
		loop of Algorithm~\ref{alg:trsvi} and $M$ steps to compute $\bm{\phi}_n^*$. For the corresponding natural gradient ascent
		we used $m M / 2$ steps to compute $\bm{\phi}_n^*$. The two parameters were selected from $(m, M) \in \{ (5, 40), (10, 20), (20, 10), (5, 100) \}$.
		We ran the trust-region method for 3 epochs and natural gradient ascent for 6 epochs. This gave both methods roughly the same amount of
		running time. The initial prior parameters were $\eta \in \{ 0.01, 0.05, 0.1, 0.2, 0.3 \}$ and $\alpha_k \in \{0.01, 0.05, 0.1\}$ but
		were updated via empirical Bayes.

		The two methods were evaluated on a separate validation set of 10,000 Wikipedia articles.
		We computed the expected value of $\bm{\beta}$ under the variational distribution and estimated the log-likelihood of the data
                given $\bm{\beta}$ using
		the Chib-style estimator described by \citet{Wallach:2009}. The trust-region
		method consistently outperformed natural gradient ascent. For 100 topics, the performance of natural
		gradient ascent varies widely while the trust-region method gave good results in almost all runs.
		In Table~\ref{tbl:wikipedia} we highlight the hyperparameters of some of the simulations.
		The table suggests that unlike natural gradient ascent, the trust-region method
		is able to perform well in particular with small batch sizes of as few as 50 data points.

		We next tested our streaming variant of SVI by simulating a data stream as follows.
		At each iteration of the simulation we took $R$ documents from the Wikipedia dataset and added
		it to a set of \textit{active documents}, where $R$ was sampled from a Poisson distribution with a fixed rate.
		We then observed one word of each active document with probability $p$ and added it to a database.
		This means that at any point in time, the database contained a number of complete and incomplete articles.
		Streaming SVI only had access to this database. To simulate the gradual accumulation of data
		we added a small delay after each observation.  
		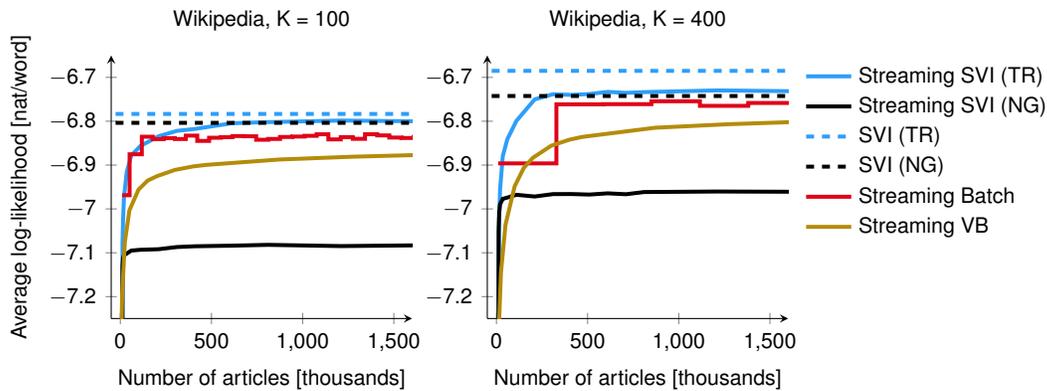
\begin{figure*}[t]
			\centering
			\begin{tikzpicture}
				\begin{axis}[
					scale only axis,
					width=4cm,
					height=3.5cm,
					at={(5.0cm, 0.0cm)},
					title={{Wikipedia, K = 400}},
					xmin=-50,
					xmax=1600,
					ymin=-7.25,
					ymax=-6.65,
					xlabel={Number of articles [thousands]},
					legend entries={Streaming SVI (TR),Streaming SVI (NG),SVI (TR),SVI (NG),Streaming Batch,Streaming VB},
					legend cell align=left,
					legend style={
						at={(1.025,1.0)},
						anchor=north west,
						font=\fontsize{8pt}{8pt}\sansmath\sffamily\selectfont,
						draw=none
					},
					xlabel near ticks,
					ytick={-7.6,-7.5,-7.4,-7.3,-7.2,-7.1,-7.0,-6.9,-6.8,-6.7,-6.6},
					xticklabel={\pgfmathprintnumber[precision=4]{\tick}},
					yticklabel={\pgfmathprintnumber[precision=4]{\tick}},
					zticklabel={\pgfmathprintnumber[precision=4]{\tick}},
					title style={font=\fontsize{8pt}{8pt}\sansmath\sffamily\selectfont},
					label style={font=\fontsize{8pt}{8pt}\sansmath\sffamily\selectfont},
					tick label style={font=\fontsize{8pt}{8pt}\sansmath\sffamily\selectfont},
					axis x line=bottom,
					axis y line=left]
					\addplot+[
						no marks,
						solid,
						line width=1.5pt,
						color={rgb,255:red,45;green,158;blue,250},
						opacity=1.0,
						mark options={solid, fill opacity=1.0}] coordinates {
						(10, -7.955518)
						(12, -7.163412)
						(15, -7.028904)
						(20, -6.955575)
						(35, -6.879488)
						(60, -6.840981)
						(110, -6.799091)
						(210, -6.750142)
						(310, -6.738413)
						(410, -6.740016)
						(510, -6.737919)
						(610, -6.733008)
						(710, -6.735393)
						(810, -6.733572)
						(1210, -6.729725)
						(1610, -6.731486)
						(2410, -6.725796)
						(3210, -6.730787)
					};
					\addplot+[
						no marks,
						solid,
						line width=1.5pt,
						color={rgb,255:red,0;green,0;blue,0},
						opacity=1.0,
						mark options={solid, fill opacity=1.0}] coordinates {
						(10, -8.107932)
						(12, -7.050207)
						(15, -7.008372)
						(20, -6.989429)
						(35, -6.976977)
						(60, -6.974092)
						(110, -6.967773)
						(210, -6.971702)
						(310, -6.96622)
						(410, -6.966043)
						(510, -6.967263)
						(610, -6.964436)
						(710, -6.966508)
						(810, -6.961498)
						(1210, -6.960305)
						(1610, -6.960836)
						(2410, -6.96461)
						(3210, -6.958625)
					};
					\addplot+[
						no marks,
						dashed,
						line width=1.5pt,
						color={rgb,255:red,45;green,158;blue,250},
						opacity=1.0,
						mark options={solid, fill opacity=1.0}] coordinates {
						(-200, -6.685097)
						(3000, -6.685097)
					};
					\addplot+[
						no marks,
						dashed,
						line width=1.5pt,
						color={rgb,255:red,0;green,0;blue,0},
						opacity=1.0,
						mark options={solid, fill opacity=1.0}] coordinates {
						(-200, -6.742448)
						(3000, -6.742448)
					};
					\addplot+[
						no marks,
						solid,
						line width=1.5pt,
						color={rgb,255:red,223;green,4;blue,25},
						opacity=1.0,
						mark options={solid, fill opacity=1.0},
						const plot] coordinates {
						(10, -6.896142)
						(62, -6.896142)
						(329, -6.761469)
						(585, -6.761053)
						(850, -6.754453)
						(1116, -6.764925)
						(1381, -6.758353)
						(1646, -6.761602)
						(1909, -6.760421)
						(2175, -6.756568)
						(2439, -6.756312)
						(2704, -6.76004)
						(2967, -6.764925)
						(3232, -6.764375)
						(3494, -6.765589)
						(3758, -6.753152)
						(3781, -6.75493)
						(4000, -6.756826)
					};
					\addplot+[
						no marks,
						solid,
						line width=1.5pt,
						color={rgb,255:red,180;green,140;blue,10},
						opacity=1.0,
						mark options={solid, fill opacity=1.0}] coordinates {
						(1, -7.725545)
						(5, -7.445259)
						(10, -7.308663)
						(25, -7.144645)
						(50, -7.035824)
						(100, -6.94856)
						(150, -6.906467)
						(200, -6.883432)
						(300, -6.85592)
						(400, -6.842156)
						(470, -6.835816)
						(870, -6.814865)
						(1270, -6.806850)
						(1670, -6.801269)
					};
				\end{axis}
				\begin{axis}[
					scale only axis,
					width=4cm,
					height=3.5cm,
					at={(0.0cm, 0.0cm)},
					title={{Wikipedia, K = 100}},
					xmin=-50,
					xmax=1600,
					ymin=-7.25,
					ymax=-6.65,
					xlabel={Number of articles [thousands]},
					ylabel={Average log-likelihood [nat/word]},
					xlabel near ticks,
					ylabel near ticks,
					ytick={-7.6,-7.5,-7.4,-7.3,-7.2,-7.1,-7.0,-6.9,-6.8,-6.7,-6.6},
					xticklabel={\pgfmathprintnumber[precision=4]{\tick}},
					yticklabel={\pgfmathprintnumber[precision=4]{\tick}},
					zticklabel={\pgfmathprintnumber[precision=4]{\tick}},
					title style={font=\fontsize{8pt}{8pt}\sansmath\sffamily\selectfont},
					label style={font=\fontsize{8pt}{8pt}\sansmath\sffamily\selectfont},
					tick label style={font=\fontsize{8pt}{8pt}\sansmath\sffamily\selectfont},
					axis x line=bottom,
					axis y line=left]
					\addplot+[
						no marks,
						solid,
						line width=1.5pt,
						color={rgb,255:red,45;green,158;blue,250},
						opacity=1.0,
						mark options={solid, fill opacity=1.0}] coordinates {
						(10, -8.356785)
						(12, -7.077507)
						(15, -7.026805)
						(20, -6.974101)
						(35, -6.916425)
						(60, -6.878142)
						(110, -6.856601)
						(210, -6.83422)
						(310, -6.822032)
						(410, -6.817565)
						(610, -6.804654)
						(810, -6.802021)
						(1210, -6.798497)
						(1610, -6.800101)
						(3210, -6.789411)
					};
					\addplot+[
						no marks,
						solid,
						line width=1.5pt,
						color={rgb,255:red,0;green,0;blue,0},
						opacity=1.0,
						mark options={solid, fill opacity=1.0}] coordinates {
						(10, -8.424571)
						(12, -7.130404)
						(15, -7.112411)
						(20, -7.105651)
						(35, -7.101953)
						(60, -7.095037)
						(110, -7.093225)
						(210, -7.091886)
						(310, -7.086936)
						(410, -7.085155)
						(610, -7.083821)
						(810, -7.081953)
						(1210, -7.084778)
						(1610, -7.083126)
						(3210, -7.081205)
					};
					\addplot+[
						no marks,
						dashed,
						line width=1.5pt,
						color={rgb,255:red,45;green,158;blue,250},
						opacity=1.0,
						mark options={solid, fill opacity=1.0}] coordinates {
						(-200, -6.783401)
						(3000, -6.783401)
					};
					\addplot+[
						no marks,
						dashed,
						line width=1.5pt,
						color={rgb,255:red,0;green,0;blue,0},
						opacity=1.0,
						mark options={solid, fill opacity=1.0}] coordinates {
						(-200, -6.803675)
						(3000, -6.803675)
					};
					\addplot+[
						no marks,
						solid,
						line width=1.5pt,
						color={rgb,255:red,223;green,4;blue,25},
						opacity=1.0,
						mark options={solid, fill opacity=1.0},
						const plot] coordinates {
						(10, -6.968828)
						(22, -6.968828)
						(51, -6.875377)
						(118, -6.835323)
						(178, -6.840837)
						(238, -6.838583)
						(299, -6.840583)
						(359, -6.832803)
						(422, -6.844904)
						(487, -6.837679)
						(552, -6.836426)
						(615, -6.833833)
						(680, -6.834297)
						(745, -6.842284)
						(811, -6.840412)
						(876, -6.835903)
						(943, -6.834956)
						(1010, -6.83328)
						(1077, -6.829776)
						(1141, -6.841453)
						(1209, -6.836294)
						(1275, -6.840137)
						(1342, -6.82969)
						(1409, -6.832346)
						(1475, -6.837245)
						(1542, -6.838258)
						(1609, -6.835762)
						(1676, -6.841467)
						(1742, -6.842524)
						(1809, -6.82928)
						(1876, -6.838496)
						(1943, -6.843811)
						(2009, -6.835139)
						(2076, -6.835925)
						(2142, -6.850946)
						(2209, -6.838294)
						(3143, -6.83514)
						(3781, -6.841062)
						(4000, -6.82999)
					};
					\addplot+[
						no marks,
						solid,
						line width=1.5pt,
						color={rgb,255:red,180;green,140;blue,10},
						opacity=1.0,
						mark options={solid, fill opacity=1.0}] coordinates {
						(1, -7.553561)
						(5, -7.309819)
						(10, -7.196596)
						(25, -7.075138)
						(50, -7.003042)
						(100, -6.955818)
						(150, -6.935376)
						(200, -6.925421)
						(300, -6.91098)
						(400, -6.903021)
						(470, -6.899616)
						(870, -6.887387)
						(1270, -6.880983)
						(1670, -6.87661)
						(2470, -6.872672)
					};
				\end{axis}
			\end{tikzpicture}
			\caption{Performance of LDA on streamed Wikipedia articles. The horizontal axis shows the number of articles
			(complete and incomplete) currently in the database (it is thus monotonically related to time).
			The vertical axis shows an estimate of the average log-likelihood on a test set separate
			from the validation set used in Figure~\ref{fig:parameter_search}. Dashed lines indicate the performance
			of SVI trained on the entire dataset and evaluated on the test set using the same hyperparameters as
			the streaming variants.}
			\label{fig:wikipedia}
		\end{figure*}

		We compared our algorithm to streaming variational Bayes \citep[SVB;][]{Broderick:2013} and the streaming batch algorithm described in
		Section~\ref{sec:streamingbatch}. For a fair comparison, we extended SVB to also learn the $\alpha_k$ via empirical Bayes.
		Since in SVB $\eta$ is only used in the first update to $\bm{\lambda}$ (Equation~\ref{eq:svb}),
		we did not estimate $\eta$ but instead ran SVB for $\eta \in \{ 0.01, 0.05, 0.1, 0.2, 0.3 \}$. The performance we report
		is the maximum performance over all $\eta$. Since SVB can only process one document (or batch of documents) at a time but
		not one word at a time, SVB was given access to the entire document when the first word was
		added to the database, putting the other algorithms at a disadvantage.
		If we naively applied SVB to LDA on a word-by-word basis, the beliefs about
		$\bm{\theta}_n$ would only be updated once after seeing the first word due to the mean-field
		approximation decoupling $\bm{\theta}_n$ from $\mathbf{x}_n$. Upon arrival of a subsequent
		word, the target distribution becomes
		\begin{align}
			\tilde p(\bm{\theta}_n, \mathbf{z}_n) \propto q(\bm{\theta}_n) q(\mathbf{z}_n) p(\mathbf{x}_n \mid \mathbf{z}_n),
		\end{align}
		so that an update will not change $q(\bm{\theta}_n)$. SVB would thus completely ignore dependencies between the
		topic assignments of different words.
		In our experiments, we used a batch size of 1,000 with SVB. The streaming batch algorithm was trained for
		100 iterations on each randomly selected set of 50,000 data points.

		We tested the streaming algorithms on LDA with 100 and 400 topics. For the hyperparameters of SVI we chose the best
		performing set of parameters found in the non-streaming setting (Figure~\ref{fig:parameter_search}A). That is, for
		the trust-region method and natural gradient ascent we used hyperparameter sets 4 and 3,
		and 1 and 2, respectively (see Table~\ref{tbl:wikipedia}). We only reduced $\kappa$ to 0.5 for 100 topics and to 0.4 for 400 topics,
		finding that keeping the learning rate up for a longer time slightly improves performance.
		$\eta$ and the $\alpha_k$ of the streaming batch algorithm
		were initialized with the same parameters used with the trust-region method.

		We find that natural gradient ascent gives extremely poor performance in the streaming setting and gets stuck
		in a local optimum early on. Using our trust-region method, on the other hand, we were able to achieve
		a performance at least as good as the best performing natural gradient ascent of the non-streaming setting
		(Figure~\ref{fig:wikipedia}).
		The simple streaming batch algorithm also performs well, although its performance is limited by the fixed batch size.
		
		SVB with a batch size of 1,000 performs much better than streaming SVI with natural gradient steps, but worse than
		our trust-region method or the streaming batch algorithm. We note that by using larger batches, e.g.
		of size 50,000, SVB can be made to perform at least as well in the limit as the streaming batch algorithm (assuming the data is stationary,
		as is the case here). However, SVB would only be able to start training once the first 50,000 data points
		have arrived. The streaming batch algorithm, on the other hand, will simply take all data points available in the
		database.

	\subsection{Netflix}
		We applied LDA to the task of collaborative filtering using the Netflix dataset of
		480,000 user ratings on 17,000 movies. Each rating is between 1 and 5. Like
		\citet{Gopalan:2013}, we thresholded the ratings and for each user kept only a list of movies with ratings of 4 or 5.
		For this ``implicit'' version of the Netflix data,
		the task is to predict which movies a user likes based on movies the user previously liked.
		20\% of each user's movies were kept for evaluation purposes and were not part of the training data.
		Again following \citet{Gopalan:2013}, we evaluate algorithms by computing \textit{normalized precision-at-M} and \textit{recall-at-M}.
		Let $L$ be the test set of movies liked by a user and let $P$ be a list of $M$ predicted movies.
		Then the two measures are given by
		\begin{align}
			\text{precision} &= \textstyle\frac{|L \cap P|}{\min\{|L|, |P|\}}, &
			\text{recall} &= \textstyle\frac{|L \cap P|}{|L|}.
		\end{align}
		Precision is normalized by $\min\{|L|, |P|\}$ rather than $|P|$ so as to
		keep users with fewer than $M$ movies from having only a small influence on the
		performance \cite{Gopalan:2013}. In the following, we will use $M = 20$.
		\begin{figure*}[t]
			\centering
			\begin{tikzpicture}
				\begin{axis}[
					scale only axis,
					width=4cm,
					height=3.5cm,
					at={(5.5cm, 0.0cm)},
					title={Netflix},
					xmin=-10,
					xmax=300,
					ymin=5,
					ymax=30,
					xlabel={Number of users [thousands]},
					ylabel={Recall [\%]},
					legend entries={SVI (TR),SVI (NG),HPF,PMF,NMF,Streaming SVI (TR),Streaming SVI (NG),Streaming Batch,Streaming VB},
					legend cell align=left,
					legend style={
						at={(1.025,1.0)},
						anchor=north west,
						font=\fontsize{8pt}{8pt}\sansmath\sffamily\selectfont,
						draw=none
					},
					xlabel near ticks,
					ylabel near ticks,
					ylabel style={at={(-0.12, 0.5)}},
					ytick={0,5,10,15,20,25,30},
					xticklabel={\pgfmathprintnumber[precision=4]{\tick}},
					yticklabel={\pgfmathprintnumber[precision=4]{\tick}},
					zticklabel={\pgfmathprintnumber[precision=4]{\tick}},
					title style={font=\fontsize{8pt}{8pt}\sansmath\sffamily\selectfont},
					tick label style={font=\fontsize{8pt}{8pt}\sansmath\sffamily\selectfont},
					label style={font=\fontsize{8pt}{8pt}\sansmath\sffamily\selectfont},
					axis x line=bottom,
					axis y line=left]
					\addplot+[
						no marks,
						dashed,
						line width=1.5pt,
						color={rgb,255:red,45;green,158;blue,250},
						opacity=1.0,
						mark options={solid, fill opacity=1.0}] coordinates {
						(-20, 20.33)
						(500, 20.33)
					};
					\addplot+[
						no marks,
						dashed,
						line width=1.5pt,
						color={rgb,255:red,0;green,0;blue,0},
						opacity=1.0,
						mark options={solid, fill opacity=1.0}] coordinates {
						(-20, 19.08)
						(500, 19.08)
					};
					\addplot+[
						no marks,
						dashed,
						line width=1.5pt,
						color={rgb,255:red,35;green,135;blue,28},
						opacity=1.0,
						mark options={solid, fill opacity=1.0}] coordinates {
						(-20, 17.735)
						(500, 17.735)
					};
					\addplot+[
						no marks,
						dashed,
						line width=1.5pt,
						color={rgb,255:red,106;green,181;blue,41},
						opacity=1.0,
						mark options={solid, fill opacity=1.0}] coordinates {
						(-20, 16.082)
						(500, 16.082)
					};
					\addplot+[
						no marks,
						dashed,
						line width=1.5pt,
						color={rgb,255:red,160;green,231;blue,54},
						opacity=1.0,
						mark options={solid, fill opacity=1.0}] coordinates {
						(-20, 6.571)
						(500, 6.571)
					};
					\addplot+[
						no marks,
						solid,
						line width=1.5pt,
						color={rgb,255:red,45;green,158;blue,250},
						opacity=1.0,
						mark options={solid, fill opacity=1.0}] coordinates {
						(10, 4.21)
						(12, 11.99)
						(15, 13.4)
						(20, 15.35)
						(35, 16.76)
						(60, 17.79)
						(110, 18.03)
						(161, 18.51)
						(210, 18.63)
						(260, 19.07)
						(310, 19.17)
					};
					\addplot+[
						no marks,
						solid,
						line width=1.5pt,
						color={rgb,255:red,0;green,0;blue,0},
						opacity=1.0,
						mark options={solid, fill opacity=1.0}] coordinates {
						(10, 2.79)
						(15, 6.11)
						(20, 6.55)
						(35, 7.69)
						(60, 7.36)
						(110, 7.77)
						(161, 7.74)
						(210, 7.76)
						(260, 7.9)
						(310, 7.49)
					};
					\addplot+[
						no marks,
						solid,
						line width=1.5pt,
						color={rgb,255:red,223;green,4;blue,25},
						opacity=1.0,
						mark options={solid, fill opacity=1.0},
						const plot] coordinates {
						(10, 8.12)
						(20, 8.12)
						(28, 13.59)
						(37, 15.78)
						(44, 15.83)
						(50, 16.7)
						(57, 16.72)
						(62, 16.18)
						(66, 16.61)
						(70, 15.77)
						(75, 15.8)
						(80, 16.69)
						(84, 16.34)
						(90, 16.77)
						(95, 16.46)
						(102, 15.64)
						(109, 16.53)
						(117, 16.48)
						(126, 16.71)
						(135, 17.16)
						(143, 16.42)
						(151, 17.2)
						(161, 16.64)
						(170, 17.21)
						(180, 17.09)
						(188, 17.35)
						(198, 16.12)
						(208, 17.9)
						(220, 17.8)
						(232, 17.52)
						(249, 17.95)
						(290, 17.0)
						(476, 16.24)
						(500, 17.84)
					};
					\addplot+[
						no marks,
						solid,
						line width=1.5pt,
						color={rgb,255:red,180;green,140;blue,10},
						opacity=1.0,
						mark options={solid, fill opacity=1.0}] coordinates {
						(1, 11.07)
						(5, 13.06)
						(10, 13.7)
						(25, 14.0)
						(50, 14.43)
						(100, 14.7)
						(150, 14.75)
						(200, 14.89)
						(300, 15.2)
						(400, 15.35)
						(470, 15.41)
					};
				\end{axis}
				\begin{axis}[
					scale only axis,
					width=4cm,
					height=3.5cm,
					at={(0.0cm, 0.0cm)},
					title={Netflix},
					xmin=-10,
					xmax=300,
					ymin=5,
					ymax=30,
					xlabel={Number of users [thousands]},
					ylabel={Precision [\%]},
					xlabel near ticks,
					ylabel near ticks,
					ytick={0,5,10,15,20,25,30},
					xticklabel={\pgfmathprintnumber[precision=4]{\tick}},
					yticklabel={\pgfmathprintnumber[precision=4]{\tick}},
					zticklabel={\pgfmathprintnumber[precision=4]{\tick}},
					title style={font=\fontsize{8pt}{8pt}\sansmath\sffamily\selectfont},
					label style={font=\fontsize{8pt}{8pt}\sansmath\sffamily\selectfont},
					tick label style={font=\fontsize{8pt}{8pt}\sansmath\sffamily\selectfont},
					axis x line=bottom,
					axis y line=left]
					\addplot+[
						no marks,
						dashed,
						line width=1.5pt,
						color={rgb,255:red,45;green,158;blue,250},
						opacity=1.0,
						mark options={solid, fill opacity=1.0}] coordinates {
						(-20, 27.48)
						(500, 27.48)
					};
					\addplot+[
						no marks,
						dashed,
						line width=1.5pt,
						color={rgb,255:red,0;green,0;blue,0},
						opacity=1.0,
						mark options={solid, fill opacity=1.0}] coordinates {
						(-20, 25.94)
						(500, 25.94)
					};
					\addplot+[
						no marks,
						dashed,
						line width=1.5pt,
						color={rgb,255:red,35;green,135;blue,28},
						opacity=1.0,
						mark options={solid, fill opacity=1.0}] coordinates {
						(-20, 24.331)
						(500, 24.331)
					};
					\addplot+[
						no marks,
						dashed,
						line width=1.5pt,
						color={rgb,255:red,106;green,181;blue,41},
						opacity=1.0,
						mark options={solid, fill opacity=1.0}] coordinates {
						(-20, 21.548)
						(500, 21.548)
					};
					\addplot+[
						no marks,
						dashed,
						line width=1.5pt,
						color={rgb,255:red,160;green,231;blue,54},
						opacity=1.0,
						mark options={solid, fill opacity=1.0}] coordinates {
						(-20, 9.163)
						(500, 9.163)
					};
					\addplot+[
						no marks,
						solid,
						line width=1.5pt,
						color={rgb,255:red,45;green,158;blue,250},
						opacity=1.0,
						mark options={solid, fill opacity=1.0}] coordinates {
						(10, 6.07)
						(12, 15.98)
						(15, 18.04)
						(20, 20.6)
						(35, 22.91)
						(60, 24.46)
						(110, 25.0)
						(161, 25.58)
						(210, 25.77)
						(260, 26.25)
						(310, 26.36)
					};
					\addplot+[
						no marks,
						solid,
						line width=1.5pt,
						color={rgb,255:red,0;green,0;blue,0},
						opacity=1.0,
						mark options={solid, fill opacity=1.0}] coordinates {
						(10, 4.48)
						(15, 8.85)
						(20, 9.56)
						(35, 11.3)
						(60, 11.25)
						(110, 11.83)
						(161, 11.75)
						(210, 11.85)
						(260, 12.17)
						(310, 11.8)
					};
					\addplot+[
						no marks,
						solid,
						line width=1.5pt,
						color={rgb,255:red,223;green,4;blue,25},
						opacity=1.0,
						mark options={solid, fill opacity=1.0},
						const plot] coordinates {
						(10, 11.35)
						(20, 11.35)
						(28, 18.46)
						(37, 21.47)
						(44, 21.61)
						(50, 22.82)
						(57, 22.87)
						(62, 22.35)
						(66, 22.85)
						(70, 21.96)
						(75, 21.98)
						(80, 23.2)
						(84, 22.8)
						(90, 23.35)
						(95, 23.02)
						(102, 22.16)
						(109, 23.23)
						(117, 22.85)
						(126, 23.53)
						(135, 23.84)
						(143, 23.25)
						(151, 23.69)
						(161, 23.08)
						(170, 23.73)
						(180, 23.83)
						(188, 24.07)
						(198, 22.88)
						(208, 24.53)
						(220, 24.39)
						(232, 24.31)
						(249, 24.54)
						(290, 23.69)
						(476, 23.06)
						(500, 24.42)
					};
					\addplot+[
						no marks,
						solid,
						line width=1.5pt,
						color={rgb,255:red,180;green,140;blue,10},
						opacity=1.0,
						mark options={solid, fill opacity=1.0}] coordinates {
						(1, 16.37)
						(5, 18.7)
						(10, 19.44)
						(25, 19.82)
						(50, 20.34)
						(100, 20.65)
						(150, 20.68)
						(200, 20.85)
						(300, 21.16)
						(400, 21.26)
						(470, 21.29)
					};
				\end{axis}
			\end{tikzpicture}
			\caption{Performance of LDA trained on a stream of Netflix users. The horizontal axis
				gives the current number of users in the database.}
			\label{fig:netflix}
		\end{figure*}
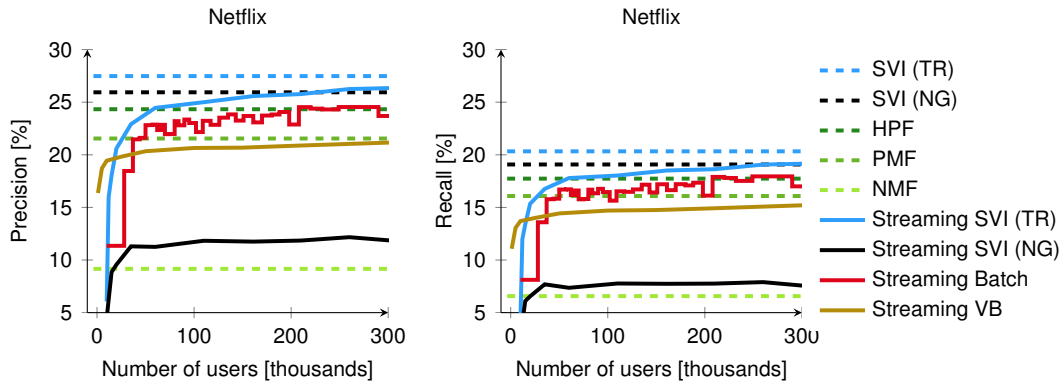

		\begin{table}[t]
			\caption{Hyperparameters corresponding to Wikipedia experiments
			in Figure~\ref{fig:parameter_search}A. $B$ is the batch size and $m$ and $M$
			control the number of iterations in the inner loops of each trust-region step
			(see text for details).}
			\label{tbl:wikipedia}
			\vskip 0.15in
			\begin{center}
			\begin{small}
			\begin{sc}
				\begin{tabular}{lcccccccr}
					\hline
					\abovespace
					\belowspace
					\# & $B$ & $\alpha$ & $\eta$ & $\kappa$ & $\tau$ & $m$ & $M$ \\
					\hline
					\abovespace
					1 &  500 & 0.1 & 0.2  & 0.6 &  10 & 10 &  20 \\
					2 & 1000 & 0.1 & 0.05 & 0.7 & 100 & 20 &  10 \\
					3 &  500 & 0.1 & 0.2  & 0.7 & 100 & 10 &  20 \\
					4 &   50 & 0.1 & 0.2  & 0.7 &  10 & 20 &  10 \\
					5 &   50 & 0.1 & 0.2  & 0.7 &   1 & 20 &  10 \\
					\belowspace
					6 &   10 & 0.1 & 0.01 & 0.5 & 100 & 10 &  20 \\
					\hline
				\end{tabular}
			\end{sc}
			\end{small}
			\end{center}
			\vskip -0.1in
		\end{table}
		\begin{table}[t]
			\caption{Hyperparameters corresponding to the highlighted Netflix experiments
			in Figure~\ref{fig:parameter_search}B.}
			\label{tbl:netflix}
			\vskip 0.15in
			\begin{center}
			\begin{small}
			\begin{sc}
				\begin{tabular}{lcccccccr}
					\hline
					\abovespace
					\belowspace
					\# & $B$ & $\alpha$ & $\eta$ & $\kappa$ & $\tau$ & $m$ & $M$ \\
					\hline
					\abovespace
					1 & 10  & 0.1  & 0.05 & 1.0 & 1   & 5  & 40 \\
					2 & 100 & 0.05 & 0.3  & 0.5 & 100 & 5  & 40 \\
					\belowspace
					3 & 500 & 0.1  & 0.3  & 0.5 & 10  & 10 & 20 \\
					\hline
				\end{tabular}
			\end{sc}
			\end{small}
			\end{center}
			\vskip -0.1in
		\end{table}
		LDA is trained on the Netflix data as in the case of Wikipedia articles, only that here users play
		the role of articles and movies play the role of words. For prediction, we approximated the
		predictive distribution over the next movie liked by user $n$,
		\begin{align}
			\textstyle
			p(x_{ni} \mid \mathbf{x}_{n,<i}) \approx \sum_k \mathbb{E}_q[\theta_{nk}] \mathbb{E}_q[\beta_{kx_{ni}}],
		\end{align}
		where $\mathbf{x}_{n,<i}$ corresponds to the 80\% movies not used for evaluation. We then took the 20 most
		likely movies under this predictive distribution which were not already in $\mathbf{x}_{n,<i}$ as predictions.

		Figure~\ref{fig:parameter_search}B shows results obtained with random hyperparameters using the same grid used
		with the Wikipedia dataset. Evaluation was performed on a random subset of 10,000 users. Interpretation of the
		results here is a little bit more challenging since we have to take into account two competing measures. Nevertheless,
		we find that for the better performing hyperparameter settings, the trust-region method again consistently outperforms
		natural gradient ascent on both measures.
		Hyperparameters of the highlighted simulations are given in Table~\ref{tbl:netflix}.

		The streamed Wikipedia articles allowed us to study the effect of a growing dataset on SVI
		without having to worry about nonstationarities in the data. Here we study a slightly more realistic
		setting by streaming the users in the order they signed up for Netflix, thus potentially creating
		a nonstationary data stream. We tested LDA only with 100 topics since using more topics did not seem to
		help improve performance (Figure~\ref{fig:parameter_search}B). Figure~\ref{fig:netflix} shows
		the results of the streaming experiment on a test set of 10,000 users different from the ones used in the
		parameter search. The hyperparameters were again chosen based on the parameter search results (sets 2 and
		3 in Table~\ref{tbl:netflix}) but reducing $\kappa$ to 0.4. As in the Wikipedia experiment, we
		find that natural gradient ascent performs poorly while the trust-region method is able to get to
		the same performance level as the best performing natural gradient ascent in the non-streaming setting.

		For comparison, we also include results of \citet{Gopalan:2013} on hierarchical Poisson
		factorization (HPF), probabilistic matrix factorization (PMF), and nonnegative matrix
		factorization (NMF) to show that our numbers are competitive\footnotemark.

\section{Discussion}
	We have proposed a new variant of stochastic variational inference which replaces the natural gradient
	step with a trust-region step. Our experiments show that this change can make SVI less prone to
	local optima and less sensitive to the choice of hyperparameters. We only explored an application
	of the trust-region method to mean-field approximations with conjugate priors. However, the same
	ideas might be applied in other settings, for example, by combining them with other
	recent innovations in variational inference such as \textit{structured SVI} \cite{Hoffman:2015}
	and \textit{black box variational inference} \cite{Ranganath:2014}.

	We further described a simple strategy for applying SVI to streaming data and have shown that the trust-region updates are crucial for good performance
	in this setting. However, Figures~\ref{fig:wikipedia} and \ref{fig:netflix} also reveal room for improvement as our streaming
	method does not yet reach the performance of the trust-region method applied to the full dataset.
	Since we used empirical Bayes to tune the parameters of the prior and the trust-region method's performance
	is not very sensitive to the batch size, the only hyperparameters still requiring some tuning are the ones
	controlling the learning rate schedule. Here we found that using larger learning rates ($\kappa \leq 0.5)$ generally
	works well. Nevertheless, it would be desirable to find a fully automatic solution. We explored adapting the work
	of \citet{Ranganath:2013} to our trust-region method but found that it did not work well in the streaming setting.

	\footnotetext{\citet{Gopalan:2013} report worse results for LDA with SVI. This may be due to a suboptimal choice of hyperparameters.}

\section*{Acknowledgments}
	This research was mostly done while Lucas Theis was an intern at Adobe Research, San Francisco.

\bibliography{references}

\begin{thebibliography}{27}
\providecommand{\natexlab}[1]{#1}
\providecommand{\url}[1]{\texttt{#1}}
\expandafter\ifx\csname urlstyle\endcsname\relax
  \providecommand{\doi}[1]{doi: #1}\else
  \providecommand{\doi}{doi: \begingroup \urlstyle{rm}\Url}\fi

\bibitem[Beck \& Teboulle(2003)Beck and Teboulle]{Beck:2003}
Beck, A. and Teboulle, M.
\newblock Mirror descent and nonlinear projected subgradient methods for convex
  optimization.
\newblock \emph{Operations Research Letters}, 31:\penalty0 167--175, 2003.

\bibitem[Blei et~al.(2003)Blei, Ng, and Jordan]{Blei:2003}
Blei, D.~M., Ng, A.~Y., and Jordan, M.~I.
\newblock {Latent Dirichlet Allocation}.
\newblock \emph{Journal of Machine Learning Research}, 3:\penalty0 993--1022,
  2003.

\bibitem[Bottou(1998)]{Bottou:1998}
Bottou, L.
\newblock \emph{{Online Learning and Stochastic Approximations}}, volume~17,
  pp.\  9--42.
\newblock Cambridge University Press, 1998.

\bibitem[Broderick et~al.(2013)Broderick, Boyd, Wibisono, Wilson, and
  Jordan]{Broderick:2013}
Broderick, T., Boyd, N., Wibisono, A., Wilson, A.~C., and Jordan, M.~I.
\newblock {Streaming Variational Bayes}.
\newblock In \emph{Advances in Neural Information Processing Systems 26}, 2013.

\bibitem[Gopalan et~al.(2013)Gopalan, Hofman, and Blei]{Gopalan:2013}
Gopalan, P., Hofman, J.~M., and Blei, D.~M.
\newblock {Scalable Recommendation with Poisson Factorization}.
\newblock \emph{arXiv.org}, abs/1311.1704, 2013.

\bibitem[Gopalan et~al.(2014)Gopalan, Charlin, and Blei]{Gopalan:2014}
Gopalan, P., Charlin, L., and Blei, D.~M.
\newblock {Content-based recommendations with Poisson factorization}.
\newblock In \emph{Advances in Neural Information Processing Systems 27}, 2014.

\bibitem[Hoffman \& Blei(2014)Hoffman and Blei]{Hoffman:2015}
Hoffman, M.~D. and Blei, D.~M.
\newblock {Structured Stochastic Variational Inference}.
\newblock \emph{arXiv.org}, abs/1404.4114, 2014.

\bibitem[Hoffman et~al.(2010)Hoffman, Blei, and Bach]{Hoffman:2010}
Hoffman, M.~D., Blei, D.~M., and Bach, F.~R.
\newblock {Online Learning for Latent Dirichlet Allocation}.
\newblock In \emph{Advances in Neural Information Processing Systems 23}, 2010.

\bibitem[Hoffman et~al.(2013)Hoffman, Blei, Wang, and Paisley]{Hoffman:2013}
Hoffman, M.~D., Blei, D.~M., Wang, C., and Paisley, J.
\newblock {Stochastic Variational Inference}.
\newblock \emph{Journal of Machine Learning Research}, 14:\penalty0 1303--1347,
  2013.

\bibitem[Hughes \& Sudderth(2013)Hughes and Sudderth]{Hughes:2013}
Hughes, M.~C. and Sudderth, E.~B.
\newblock {Memoized Online Variational Inference for Dirichlet Process Mixture
  Models}.
\newblock In \emph{Advances in Neural Information Processing Systems 26}, 2013.

\bibitem[Lin et~al.(2007)Lin, Weng, and Keerthi]{Lin:2007}
Lin, C.-J., Weng, R.~C., and Keerthi, S.~S.
\newblock Trust region newton methods for large-scale logistic regression.
\newblock In \emph{Proceedings of the 26th International Conference on Machine
  Learning}, 2007.

\bibitem[Mandt \& Blei(2014)Mandt and Blei]{Mandt:2014}
Mandt, S. and Blei, D.
\newblock {Smoothed Gradients for Stochastic Variational Inference}.
\newblock In \emph{Advances in Neural Information Processing Systems 27}, 2014.

\bibitem[Maybeck(1982)]{Maybeck:1982}
Maybeck, P.~S.
\newblock \emph{{Stochastic models, estimation and control}}.
\newblock Academic Press, 1982.

\bibitem[Minka(2001)]{Minka:2001}
Minka, T.
\newblock \emph{{A family of algorithms for approximate Bayesian inference}}.
\newblock PhD thesis, MIT Media Lab, MIT, 2001.

\bibitem[Nemirovski \& Yudin(1983)Nemirovski and Yudin]{Nemirovski:1983}
Nemirovski, A. and Yudin, D.
\newblock \emph{Problem Complexity and Method Efficiency in Optimization}.
\newblock Wiley, New York, 1983.

\bibitem[Nocedal \& Wright(1999)Nocedal and Wright]{Nocedal:1999}
Nocedal, J. and Wright, S.~J.
\newblock \emph{Numerical Optimization}.
\newblock Springer, New York, 1999.

\bibitem[Paisley et~al.(2012)Paisley, Wang, and Blei]{Paisley:2012}
Paisley, J., Wang, C., and Blei, D.~M.
\newblock {The discrete infinite logistic normal distribution}.
\newblock \emph{Bayesian Analysis}, 7\penalty0 (2):\penalty0 235--272, 2012.

\bibitem[Pascanu et~al.(2014)Pascanu, Dauphin, Ganguli, and
  Bengio]{Pascanu:2014}
Pascanu, R., Dauphin, Y.~N., Ganguli, S., and Bengio, Y.
\newblock On the saddle point problem for non-convex optimization.
\newblock \emph{arXiv.org}, abs/1405.4604, 2014.

\bibitem[Ranganath et~al.(2013)Ranganath, Wang, Blei, and Xing]{Ranganath:2013}
Ranganath, R., Wang, C., Blei, D.~M., and Xing, E.~P.
\newblock {An Adaptive Learning Rate for Stochastic Variational Inference}.
\newblock In \emph{Proceedings of the 30th International Conference on Machine
  Learning}, 2013.

\bibitem[Ranganath et~al.(2014)Ranganath, Gerrish, and Blei]{Ranganath:2014}
Ranganath, R., Gerrish, S., and Blei, D.~M.
\newblock Black box variational inference.
\newblock In \emph{International Conference on Artificial Intelligence and
  Statistics 17}, 2014.

\bibitem[Sato(2001)]{Sato:2001}
Sato, M.
\newblock Online model selection based on the variational bayes.
\newblock \emph{Neural Computation}, 13\penalty0 (7):\penalty0 1649--1681,
  2001.

\bibitem[Tank et~al.(2014)Tank, Foti, and Fox]{Tank:2014}
Tank, A., Foti, N.~J., and Fox, E.~B.
\newblock {Streaming Variational Inference for Bayesian Nonparametric Mixture
  Models}.
\newblock \emph{arXiv.org}, abs/1412.0694, 2014.

\bibitem[van Hateren \& van~der Schaaf(1998)van Hateren and van~der
  Schaaf]{vanHateren:1998}
van Hateren, J.~H. and van~der Schaaf, A.
\newblock Independent component filters of natural images compared with simple
  cells in primary visual cortex.
\newblock \emph{Proc. of the Royal Society B: Biological Sciences},
  265\penalty0 (1394), 1998.

\bibitem[Wainwright \& Jordan(2008)Wainwright and Jordan]{Wainwright:2008}
Wainwright, M.~J. and Jordan, M.~I.
\newblock {Graphical Models, Exponential Families, and Variational Inference}.
\newblock \emph{Foundations and Trends in Machine Learning}, 1\penalty0 (1),
  2008.

\bibitem[Wallach et~al.(2009)Wallach, Murray, Salakhutdinov, and
  Mimno]{Wallach:2009}
Wallach, H.~M., Murray, I., Salakhutdinov, R., and Mimno, D.
\newblock {Evaluation Methods for Topic Models}.
\newblock In \emph{Proceedings of the 26th International Conference on Machine
  Learning}, 2009.

\bibitem[Wang et~al.(2011)Wang, Paisley, and Blei]{Wang:2011}
Wang, C., Paisley, J., and Blei, D.
\newblock {Online variational inference for the hierarchical Dirichlet
  process}.
\newblock In \emph{International Conference on Artificial Intelligence and
  Statistics 14}, 2011.

\bibitem[Zoran \& Weiss(2011)Zoran and Weiss]{Zoran:2011}
Zoran, D. and Weiss, Y.
\newblock {From learning models of natural image patches to whole image
  restoration}.
\newblock In \emph{International Conference on Computer Vision}, pp.\
  479--486, 2011.

\end{thebibliography}


\begin{thebibliography}{3}
\providecommand{\natexlab}[1]{#1}
\providecommand{\url}[1]{\texttt{#1}}
\expandafter\ifx\csname urlstyle\endcsname\relax
  \providecommand{\doi}[1]{doi: #1}\else
  \providecommand{\doi}{doi: \begingroup \urlstyle{rm}\Url}\fi

\bibitem[Bishop(2006)]{Bishop:2006}
C.~M. Bishop.
\newblock \emph{Pattern Recognition and Machine Learning}.
\newblock Springer, 2006.

\bibitem[Blei et~al.(2003)Blei, Ng, and Jordan]{Blei:2003}
D.~M. Blei, A.~Y. Ng, and M.~I. Jordan.
\newblock {Latent Dirichlet Allocation}.
\newblock \emph{Journal of Machine Learning Research}, 3:\penalty0 993--1022,
  2003.

\bibitem[Petersen and Pedersen(2008)]{Petersen:2008}
K.~Petersen and M.~Pedersen.
\newblock {The Matrix Cookbook}, 2008.

\end{thebibliography}
\bibliographystyle{icml2015}

\end{document}


\section{Gradient of Kullback-Leibler divergence}
		Let $\bm{\lambda}$ and $\bm{\lambda'}$ be two sets of natural parameters of an exponential
		family, that is,
		\begin{align}
			q(\bm{\beta}; \bm{\lambda}) &= h(\bm{\beta}) \exp\left( \bm{\lambda}^\top t(\bm{\beta}) - a(\bm{\lambda}) \right).
		\end{align}
		The partial derivatives of their Kullback-Leibler divergence are given by
		\begin{align}
			\frac{\partial}{\partial \bm{\lambda}} D_\text{KL}(\bm{\lambda}, \bm{\lambda}')
			&= \frac{\partial}{\partial \bm{\lambda}} \mathbb{E}_{\bm{\lambda}}\left[ \log \frac{q(\bm{\beta}; \bm{\lambda})}{q(\bm{\beta}; \bm{\lambda}')} \right] \\
			&= \frac{\partial}{\partial \bm{\lambda}} \mathbb{E}_{\bm{\lambda}}\left[ (\bm{\lambda} - \bm{\lambda}')^\top t(\bm{\beta}) - a(\bm{\lambda}) + a(\bm{\lambda}') \right] \\
			&= (\bm{\lambda} - \bm{\lambda}')^\top \frac{\partial}{\partial \bm{\lambda}} \mathbb{E}_{\bm{\lambda}}\left[ t(\bm{\beta}) \right] + \mathbb{E}_{\bm{\lambda}}\left[ t(\bm{\beta}) \right] - \frac{\partial}{\partial \bm{\lambda}} a(\bm{\lambda}) \\
			&= (\bm{\lambda} - \bm{\lambda}')^\top \frac{\partial^2}{\partial \bm{\lambda}^2} a(\bm{\lambda}) + \frac{\partial}{\partial \bm{\lambda}} a(\bm{\lambda}) - \frac{\partial}{\partial \bm{\lambda}} a(\bm{\lambda}) \\
			&= (\bm{\lambda} - \bm{\lambda}')^\top I(\bm{\lambda}),
		\end{align}
		where we have used the exponential family identities
		\begin{align}
			\frac{\partial}{\partial \bm{\lambda}} a(\bm{\lambda}) &= \mathbb{E}_{\bm{\lambda}}[t(\bm{\beta})]^\top,  &
			\frac{\partial^2}{\partial \bm{\lambda}^2} a(\bm{\lambda}) &= I(\bm{\lambda}).
		\end{align}

	\section{Asymptotic behavior of trust-region method}
		Here we show that the trust-region update given below converges to a natural gradient step.
		\begin{align}
			d\bm{\lambda} = \argmax{d\bm{\lambda}} \left\{ \mathcal{L}_n(\bm{\lambda} + d\bm{\lambda})
				- \xi D_\text{KL}(\bm{\lambda} + d\bm{\lambda}, \bm{\lambda}) \right\}
		\end{align}
		For large $\xi$, $d\bm{\lambda}$ will be close to zero so that we can focus on the target
		functions' first-order terms. For exponential families in canonical form, we have
		\begin{align}
			D_\text{KL}(\bm{\lambda} + d\bm{\lambda}, \bm{\lambda})
			&= E_{\bm{\lambda} + d\bm{\lambda}}\left[ \log \frac{q(\bm{\beta}; \bm{\lambda} + d\bm{\lambda})}{q(\bm{\beta}; \bm{\lambda})} \right] \\
			&= E_{\bm{\lambda} + d\bm{\lambda}}\left[ d\bm{\lambda}^\top t(\bm{\beta}) - a(\bm{\lambda} + d\bm{\lambda}) + a(\bm{\lambda}) \right] \\
			&= d\bm{\lambda}^\top \nabla a(\bm{\lambda} + d\bm{\lambda}) - a(\bm{\lambda} + d\bm{\lambda}) + a(\bm{\lambda}).
		\end{align}
		The gradient of the KL divergence in $d\bm{\lambda}$ is thus given by
		\begin{align}
			\nabla D_\text{KL}(\bm{\lambda} + d\bm{\lambda}, \bm{\lambda})
			&= \nabla a(\bm{\lambda} + d\bm{\lambda}) + \nabla^2 a(\bm{\lambda} + d\bm{\lambda}) d\bm{\lambda} - \nabla a(\bm{\lambda} + d\bm{\lambda}) \\
			&= \nabla^2 a(\bm{\lambda} + d\bm{\lambda}) d\bm{\lambda}.
		\end{align}
		Approximating the target function around $\bm{\lambda}$ yields
		\begin{align}
			d\bm{\lambda}^\top \nabla \mathcal{L}_n(\bm{\lambda}) - \xi d\bm{\lambda}^\top \nabla^2 a(\bm{\lambda}) d\bm{\lambda},
		\end{align}
		which when maximized gives an update proportional to the natural gradient direction,
		\begin{align}
			d\bm{\lambda} = \frac{1}{2\xi} \left( \nabla^2 a(\bm{\lambda}) \right)^{-1} \nabla \mathcal{L}_n(\bm{\lambda}).
		\end{align}

	\section{Latent Dirichlet Allocation}
		In LDA we have global parameters $\bm{\beta}$ consisting of distributions over words
		$\bm{\beta}_k$ and local parameters $\bm{\theta}_n, \mathbf{z}_n$ with distributions
		\begin{align}
			p(\bm{\beta}) &= \prod_k \text{Dir}(\bm{\beta}_k; \bm{\eta}), \\
			p(\bm{\theta}_n) &= \text{Dir}(\bm{\theta}_n; \bm{\alpha}), \\
			p(\bm{z}_n \mid \bm{\theta}_n) &= \prod_m \theta_{nz_{nm}}, \\
			p(\bm{x}_n \mid z_n, \bm{\beta}) &= \prod_{m} \beta_{z_{nm}x_{nm}}.
		\end{align}
		We approximate the posterior distribution over $\bm{\beta}$ with
		\begin{align}
			q(\bm{\beta}) &= \prod_k \text{Dir}(\bm{\beta}_k; \bm{\lambda}_k).
		\end{align}
		Writing the likelihood in the form of Equation~2 of the main paper gives
		\begin{align}
			p(\mathbf{x}_n, \mathbf{z}_n, \bm{\theta}_n \mid \bm{\beta})
			&= \text{Dir}(\bm{\theta}; \bm{\alpha}) \left( \prod_m \theta_{nz_{nm}} \right) \left(\prod_m  \beta_{z_{nm}x_{nm}} \right) \\
			&= h(\bm{\theta}_n, \mathbf{z}_n) \exp\left( \sum_m \log \beta_{z_{nm}x_{nm}} \right) \\
			&= h(\bm{\theta}_n, \mathbf{z}_n) \exp\left( \left< t(\bm{\beta}), f(\mathbf{z}_n, \mathbf{x}_n) \right> \right)
		\end{align}
		where $h$ encompasses all terms which do not depend on $\bm{\beta}$ and
		\begin{align}
			t(\bm{\beta}) &= \log \bm{\beta}, \\
			f(\mathbf{z}_n, \mathbf{x}_n) &= \sum_m \mathbf{I}_{z_{nm}x_{nm}},
		\end{align}
		where $\mathbf{I}_{ij}$ is a matrix with entry $(i, j)$ set to $1$ and all other entries set
		to $0$.
		We here assume that $\bm{\beta}$ is a matrix whose rows are the topics $\bm{\beta}_k$ and
		\begin{align}
			\left< \mathbf{A}, \mathbf{B} \right> 
			= \text{vec}(\mathbf{A})^\top \text{vec}(\mathbf{B}) 
			= \text{tr}(\mathbf{A}\mathbf{B})
		\end{align}
		for matrices $\mathbf{A}$ and $\mathbf{B}$. Since the standard parametrization of the
		Dirichlet distribution is already in canonical form, we can immediately apply Equation~10 of the
		main paper to get the steps of the inner loop of the trust-region update,
		\begin{align}
			\bm{\lambda} = (1 - \rho_t) \bm{\lambda}_t + \rho_t \left( \bm{\eta} + N \mathbb{E}_{\bm{\phi}_n^*}[f(\mathbf{x}_n, \mathbf{z}_n)] \right).
		\end{align}
		The beliefs over $\mathbf{z}_n$ (i.e., $\bm{\phi}_n^*$) are computed in the usual manner \citep{Blei:2003}.
		
	\section{Mixture models}
		Consider a mixture model where the local parameters are the cluster assignments $k_n$ and the
		global parameters are the prior weights $\bm{\pi}$ and the components' parameters $\bm{\beta}_k$.
		We assume the following more specific form for the model,
		\begin{align}
			p(\bm{\pi}) &= \text{Dir}(\bm{\pi}; \bm{\alpha}), \\
			p(k_n \mid \bm{\pi}) &= \pi_{k_n}, \\
			p(\bm{\beta}_k) &\propto h(\bm{\beta}_k) \exp\left( \bm{\eta}^\top t(\bm{\beta}_k) \right), \\
			p(\mathbf{x}_n \mid k_n, \bm{\beta}) &= g(\mathbf{x}_n) \exp\left( t(\bm{\beta}_{k_n})^\top f(\mathbf{x}_n) \right)
		\end{align}
		for suitable funtions $t$, $f$, $g$, $h$.
		The factors of a mean-field approximation to the posterior are given by
		\begin{align}
			q(\bm{\pi}) &= \text{Dir}(\bm{\pi}; \bm{\gamma}), \\
			q(k_n) &= \phi_{nk_n}, \\
			q(\bm{\beta}_k) &\propto h(\bm{\beta}_k) \exp\left(\bm{\lambda}_k^\top t(\bm{\beta}_k)\right).
		\end{align}
		In each iteration, the trust-region method alternates between computing the optimal $\bm{\phi}_n$ and updating
		$\bm{\gamma}$ and $\bm{\lambda}$. The approximate posterior over mixture components,
		slightly abusing notation, is given by
		\begin{align}
			\bm{\phi}_n^*
			&= \underset{\bm{\phi}_n}{\text{argmax}} \, \mathcal{L}_n(\bm{\lambda}, \bm{\gamma}, \bm{\phi}_n) \\
			&\propto \exp \mathbb{E}_q\left[ \log p(\mathbf{x}_n \mid k_n, \bm{\beta}) p(k_n \mid \bm{\pi}) \right] \\
			&= \exp \mathbb{E}_q\left[ \log p(\mathbf{x}_n \mid k_n, \bm{\beta}) \right] \exp \left( \psi(\bm{\gamma}) - \psi\left(\sum_k \gamma_k\right) \right)
		\end{align}
		where for the expected log-likelihood we have
		\begin{align}
			\label{eq:expected_loglikelihood}
			\mathbb{E}_q\left[ \log p(\mathbf{x}_n \mid k_n, \bm{\beta}) \right]
			= \mathbb{E}_q\left[ t(\bm{\beta}_{k_n})\right]^\top f(\mathbf{x}_n) + \log g(\mathbf{x}_n).
		\end{align}
		Once $\bm{\phi}_n^*$ is computed, we update $\bm{\gamma}$ and $\bm{\lambda}$ via
		\begin{align}
			\bm{\gamma} &= (1 - \rho_t) \bm{\gamma}_t + \rho_t \left( \bm{\alpha} + N\bm{\phi}_n^* \right), \\
			\bm{\lambda}_k &= (1 - \rho_t) \bm{\lambda}_k^t + \rho_t \left( \bm{\eta} + N \phi^*_{nk} f(\mathbf{x}_n) \right).
		\end{align}
		For mini-batches of size $B$, these updates become
		\begin{align}
			\bm{\gamma} &= (1 - \rho_t) \bm{\gamma}_t + \rho_t \left(\bm{\alpha} + \frac{N}{B}\sum_n\bm{\phi}_n^*\right), \\
			\label{eq:update_component}
			\bm{\lambda}_k &= (1 - \rho_t) \bm{\lambda}_k^t + \rho_t \left( \bm{\eta} + \frac{N}{B} \sum_n \phi^*_{nk} f(\mathbf{x}_n) \right).
		\end{align}
		To use these results with any concrete mixture model, we have to write down the prior over
		$\bm{\beta}$ in canonical form and implement the expected log-likelihood (Equation~\ref{eq:expected_loglikelihood}) and the update in
		Equation~\ref{eq:update_component}.

		\subsection{Mixture of multivariate Bernoullis}
			For the multivariate Bernoulli model,
			\begin{align}
				p(x_n \mid k_n, \bm{\beta}) = \prod_i \beta_{ki}^{x_{ni}} (1 - \beta_{ki})^{1 - x_{ni}},
			\end{align}
			we assume beta distributions for the prior and approximate posterior,
			\begin{align}
				p(\bm{\beta}_k) &\propto \beta_{ki}^{a - 1}(1 - \beta_{ki})^{b - 1}, \\
				q(\bm{\beta}_k) &\propto \beta_{ki}^{a_{ki} - 1}(1 - \beta_{ki})^{b_{ki} - 1}.
			\end{align}
			$\bm{\lambda}_k = (\mathbf{a}_k, \mathbf{b}_k)$ are already the natural parameters of the beta distribution,
			so that the updates (Equation~\ref{eq:update_component}) become
			\begin{align}
				\mathbf{a}_k &= (1 - \rho_t) \mathbf{a}_k^t + \rho_t \left(a + \frac{N}{B} \sum_n \phi_{nk}^* x_{ni}\right), \\
				\mathbf{b}_k &= (1 - \rho_t) \mathbf{b}_k^t + \rho_t \left(b + \frac{N}{B} \sum_n \phi_{nk}^* (1 - x_{ni})\right).
			\end{align}
			The expected log-likelihood needed for the computation of $\bm{\phi}_n^*$ is given by
			\begin{align}
				\mathbb{E}_q\left[ \log p(\mathbf{x}_n \mid k_n, \bm{\beta}) \right]
				&= \mathbf{x}_n^\top \psi(\mathbf{a}_{k_n}) + (1 - \mathbf{x}_n)^\top \psi(\mathbf{b}_{k_n}) 
					- \bm{1}^\top \psi(\mathbf{a}_{k_n} + \mathbf{b}_{k_n}),
			\end{align}
			where the digamma function $\psi$ is applied point-wise and $\bm{1}$ is a vector of ones.

		\subsection{Mixture of Gaussians}
			We assume a normal-inverse-Wishart distribution (NIW) for the parameters $\bm{\beta}_k = (\bm{\mu}_k, \bm{\Sigma}_k)$
			of each Gaussian distribution. The NIW for a singe component is given by
			\begin{align}
				p(\bm{\mu}, \bm{\Sigma})
					&\propto \exp\left( -\frac{s}{2} (\bm{\mu} - \mathbf{m})^\top \bm{\Sigma}^{-1} (\bm{\mu} - \mathbf{m}) \right) 
						\exp\left( -\frac{1}{2} \text{tr}(\bm{\Psi}\bm{\Sigma}^{-1}) \right) |\bm{\Sigma}|^{-\frac{\nu + D + 1}{2}} \\
					&= \exp\left( -\frac{s}{2} \bm{\mu}^\top \bm{\Sigma}^{-1} \bm{\mu} + s \bm{\mu}^\top \bm{\Sigma}^{-1} \mathbf{m}
						- \frac{1}{2} \text{tr}(s\mathbf{m}\mathbf{m}^\top\bm{\Sigma}^{-1}) - \frac{1}{2} \text{tr}(\bm{\Psi}\bm{\Sigma}^{-1})
						- \frac{\nu}{2} \log |\bm{\Sigma}| - \frac{D + 1}{2} \log |\bm{\Sigma}| \right) \\
					&= \exp\left( \eta(\mathbf{m}, s, \bm{\Psi}, \nu)^\top t(\bm{\mu}, \bm{\Sigma}) \right)
			\end{align}
			where
			\begin{align}
					\eta(\mathbf{m}, s, \bm{\Psi}, \nu) 
					&= \left( s, -2s\mathbf{m}, \text{vec}(s\mathbf{m}\mathbf{m}^\top + \bm{\Psi}), \nu \right)  \\
					&= \left( s, \mathbf{b}, \text{vec}\, \mathbf{C}, \nu \right) = \bm{\eta}, \\
				t(\bm{\mu}, \bm{\Sigma})
					&= -\frac{1}{2} \left( \bm{\mu}^\top \bm{\Sigma}^{-1} \bm{\mu}, \bm{\Sigma}^{-1}\bm{\mu}, \text{vec}\, \bm{\Sigma}^{-1}, \log|\bm{\Sigma}| \right).
			\end{align}
			are the natural parameters and sufficient statistics of the distribution, respectively.
			The likelihood for a single data point $\mathbf{x}$ is given by
			\begin{align}
				p(\mathbf{x} \mid \bm{\mu}, \bm{\Sigma})
					&\propto \exp\left( -\frac{1}{2} (\mathbf{x} - \bm{\mu})^\top \bm{\Sigma}^{-1} (\mathbf{x} - \bm{\mu}) \right) / |\bm{\Sigma}|^\frac{1}{2} \\
					&= \exp\left( -\frac{1}{2} \text{tr}\left(\bm{\Sigma}^{-1} \mathbf{x}\mathbf{x}^\top \right) + \mathbf{x}^\top \bm{\Sigma}^{-1}\bm{\mu} - \frac{1}{2} \bm{\mu}^\top \bm{\Sigma}^{-1} \bm{\mu} - \frac{1}{2} \log |\bm{\Sigma}| \right) \\
					&= \exp\left( t(\bm{\mu}, \bm{\Sigma})^\top f(\mathbf{x}) \right),
			\end{align}
			where
			\begin{align}
				f(\mathbf{x}) 
					&= \left(1, -2\mathbf{x}, \text{vec}\left(\mathbf{x}\mathbf{x}^\top\right), 1 \right).
			\end{align}
			Hence, assuming the natural parameters of all components are given by
			\begin{align}
				\bm{\eta}_k 
				&= \eta(\mathbf{m}_k, s_k, \bm{\Psi}_k, \nu_k)
				= \left( s_k, \mathbf{b}_k, \text{vec}\, \mathbf{C}_k, \nu_k \right),
			\end{align}
			an update of the inner loop of the trust-region method (Equation~\ref{eq:update_component}) is given by
			\begin{align}
				s_k &= (1 - \rho_t) s_k^t + \rho_t \left( s + \frac{N}{B} \sum_n \phi_{nk}^* \right), \\
				\mathbf{b}_k &= (1 - \rho_t) \mathbf{b}_k^t + \rho_t \left( \mathbf{b} - 2 \frac{N}{B} \sum_n \phi_{nk}^* \mathbf{x}_n \right), \\
				\mathbf{C}_k &= (1 - \rho_t) \mathbf{C}_k^t + \rho_t \left(\mathbf{C} + \frac{N}{B} \sum_n \phi_{nk}^* \mathbf{x}_n\mathbf{x}_n^\top \right), \\
				\nu_k &= (1 - \rho_t) \nu_k^t + \rho_t \left(\nu + \frac{N}{B} \sum_n \phi_{nk}^* \right).
			\end{align}
			Or, in terms of the more traditional parametrization,
			\begin{align}
				\mathbf{m}_k &= (1 - \rho_t) \frac{s_k^t}{s_k} \mathbf{m}_k^t + \frac{1}{s_k} \rho_t \left( s\mathbf{m} + \frac{N}{B} \sum_n \phi_{nk}^* \mathbf{x}_n \right), \\
				\bm{\Psi}_k &= (1 - \rho_t) \left(\bm{\Psi}_k^t +
				s_k^t\mathbf{m}_k^t{\mathbf{m}_k^t}^\top\right) + \rho_t \left(\bm{\Psi} + s\mathbf{m}\mathbf{m}^\top + \frac{N}{B} \sum_n \phi_{nk}^* \mathbf{x}_n\mathbf{x}_n^\top \right) - s_k\mathbf{m}_k\mathbf{m}_k^\top.
			\end{align}
			A proof that these updates leave $\bm{\Psi}_k$ positive definite is given below.
			
			\subsubsection*{Positive definiteness of $\bm{\Psi}_k$}
				We show positive definiteness of $\bm{\Psi}_k$ in two steps. First, we show that the constraint is fulfilled
				for $\rho_t = 0$ and $\rho_t = 1$. Second, we show that the set of valid natural parameters induced by the constraint
				is convex, implying that the constraint must be fulfilled for any linear interpolation of two natural parameters.
				
				For $\rho_t = 0$, we have $\bm{\Psi}_k = \bm{\Psi}_k^t$
				since none of the natural parameters has changed. Thus, $\bm{\Psi}_k$ is positive definite if $\bm{\Psi}_k^t$ is positive definite.
				For $\rho_t = 1$, we have
				\begin{align}
					\label{eq:psi}
					\bm{\Psi}_k
					&= \bm{\Psi} + s\mathbf{m}\mathbf{m}^\top + \frac{N}{B} \sum_n \phi_{nk}^* \mathbf{x}_n\mathbf{x}_n^\top - s_k\mathbf{m}_k\mathbf{m}_k^\top, \\
					s_k\mathbf{m}_k\mathbf{m}_k
					&= \frac{1}{s_k}\left( s\mathbf{m} + \frac{N}{B} \sum_n \phi_{nk}^* \mathbf{x}_n \right) \left( s\mathbf{m} + \frac{N}{B} \sum_n \phi_{nk}^* \mathbf{x}_n \right)^\top, \\
					s_k &= s + \frac{N}{B}\sum_n \phi_{nk}^*.
				\end{align}
				Note that $p_0 = s/s_k$ and $p_n = \frac{N}{B}\phi_{nk} / s_k$
				are positive and sum to one and therefore can be considered probabilities. Let $\mathbf{X}$ be a random variable which
				takes on value $\mathbf{m}$ with probability $p_0$ and value
				$\mathbf{x}_n$ with probability $p_n$. Then we can rewrite Equation~\ref{eq:psi} as
				\begin{align}
					\bm{\Psi}_k
					&= \bm{\Psi} + s_k\mathbb{E}_p[\mathbf{XX}^\top] -  s_k \mathbb{E}_p[\mathbf{X}]\mathbb{E}_p[\mathbf{X}]^\top 
					= \bm{\Psi} + s_k \mathbb{V}[\mathbf{X}].
				\end{align}
				Since $\bm{\Psi}$ is positive definite and the covariance matrix $\mathbb{V}[\mathbf{X}]$ is at least positive semi-definite,
				$\bm{\Psi}_k$ must be positive definite.

				We next show that the set of valid natural parameters,
				\begin{align}
					\{ (s, \mathbf{b}, \mathbf{C}, \nu) : s > 0, \nu > D - 1, \mathbf{C} - \frac{1}{4s}\mathbf{b}\mathbf{b}^\top \text{ is p. d. } \},
				\end{align}
				is convex. Not that this set is convex iff
				\begin{align}
					\{ (s, \mathbf{b}, \mathbf{C}) : s > 0, \mathbf{C} - \frac{1}{s^2}\mathbf{b}\mathbf{b}^\top \text{ is p. d. } \},
				\end{align}
				is convex. This set is convex iff
				\begin{align}
					\{ (s, \mathbf{b}, \mathbf{C}) : s > 0, \mathbf{C} - \mathbf{b}\mathbf{b}^\top \text{ is p. d. } \},
				\end{align}
				is convex, since any \textit{perspective function} preserves convexity and $P(s, \mathbf{b}, \mathbf{C}) = (s, \mathbf{b}/s, \mathbf{C})$
				is a perspective function. Finally, this set is convex iff the following set is convex,
				\begin{align}
					\Omega = \{ (\mathbf{b}, \mathbf{C}) : \mathbf{C} - \mathbf{b}\mathbf{b}^\top \text{ is p. d. } \}.
				\end{align}
				Assume $(\mathbf{b}_1, \mathbf{C}_1), (\mathbf{b}_2, \mathbf{C}_2) \in \Omega$ and let $\rho \in [0, 1]$. Then
				\begin{align}
					&\quad \rho \mathbf{C}_1 + (1 - \rho) \mathbf{C}_2 - \left(\rho \mathbf{b}_1 + (1 - \rho) \mathbf{b}_2\right) \left(\rho \mathbf{b}_1 + (1 - \rho) \mathbf{b}_2\right)^\top \\
					&= \rho \mathbf{C}_1 + (1 - \rho) \mathbf{C}_2 - \left(\rho (\mathbf{b}_1 - \mathbf{b}_2) + \mathbf{b}_2\right) \left(\mathbf{b}_1 + (1 - \rho) (\mathbf{b}_2 - \mathbf{b}_1)\right)^\top \\
					&= \rho (\mathbf{C}_1 - \mathbf{b}_1\mathbf{b}_1^\top) + (1 - \rho) (\mathbf{C}_2 - \mathbf{b}_2\mathbf{b}_2^\top)
						+ \rho (1 - \rho) (\mathbf{b}_1 - \mathbf{b}_2) (\mathbf{b}_1 - \mathbf{b}_2)^\top,
				\end{align}
				which is a sum of positive definite and semi-definite matrices and therefore positive definite.
				Hence,
				\begin{align}
					\left(\rho \mathbf{C}_1 + (1 - \rho) \mathbf{C}_2, \rho \mathbf{b}_1 + (1 - \rho) \mathbf{b}_2\right) \in \Omega
				\end{align}
				and $\Omega$ is convex.

			\subsubsection*{Expected log-likelihood}
				To compute he expected log-likelihood (Equation~\ref{eq:expected_loglikelihood}), we need
				\begin{align}
					\mathbb{E}_q\left[ \bm{\mu}_k^\top \bm{\Sigma}_k^{-1} \bm{\mu}_k \right]
					&= \mathbb{E}_q\left[ \mathbb{E}_q\left[ \bm{\mu}_k^\top \bm{\Sigma}_k^{-1} \bm{\mu}_k \mid \bm{\Sigma}_k \right] \right] \\
					\label{eq:petersen}
					&= \mathbb{E}_q\left[ \text{tr}\left(s_k^{-1}\bm{\Sigma}_k\bm{\Sigma}_{k}^{-1}\right) + \mathbf{m}_k^\top \bm{\Sigma}_k^{-1} \mathbf{m}_k\right] \\
					&= Ds_k^{-1} + \nu_k \mathbf{m}_k^\top \bm{\Psi}_k^{-1} \mathbf{m}_k, \\
					\mathbb{E}_q\left[\bm{\Sigma}_k^{-1} \bm{\mu}_k \right]^\top \mathbf{x}
					&= \nu_k \mathbf{m}_k^\top \bm{\Psi}_k^{-1} \mathbf{x}, \\
					\mathbf{x}^\top \mathbb{E}_q\left[ \bm{\Sigma}_k^{-1} \right] \mathbf{x}
					&= \nu_k \mathbf{x}^\top \bm{\Psi}_k^{-1} \mathbf{x}, \\
					\label{eq:bishop}
					\mathbb{E}_q\left[\log |\bm{\Sigma}_k^{-1}|\right]
					&= \sum_{i = 1}^D \psi\left( \frac{\nu_k + 1 - i}{2} \right) + D\log 2 + \log|\bm{\Psi}_k^{-1}|.
				\end{align}
				For Equation~\ref{eq:petersen}, see \cite{Petersen:2008}. For Equation~\ref{eq:bishop}, see \cite{Bishop:2006}.
				The expected log-likelihood is thus given by
				\begin{align}
					\mathbb{E}_q\left[ \log p(\mathbf{x} \mid k, \bm{\beta}) \right]
					&= \mathbb{E}_q\left[ -\frac{1}{2}\mathbf{x}^\top \bm{\Sigma}_k^{-1} \mathbf{x} + \mathbf{x}\bm{\Sigma}_k^{-1} \bm{\mu}_k - \frac{1}{2}\bm{\mu}_k^\top \bm{\Sigma}_k^{-1} \bm{\mu}_k + \frac{1}{2} \log |\bm{\Sigma}_k^{-1}| - \frac{D}{2}\log(2\pi)\right] \\
					&= - \frac{\nu_k}{2} \mathbf{x}^\top \bm{\Psi}_k^{-1} \mathbf{x} + \nu_k \mathbf{x}^\top \bm{\Psi}_k^{-1} \mathbf{m}_k
					- \frac{D}{2} s_k^{-1} - \frac{\nu_k}{2} \mathbf{m}_k^\top \bm{\Psi}_k^{-1} \mathbf{m}_k \\
					&\quad + \frac{1}{2} \left(\sum_{i = 1}^D \psi\left( \frac{\nu_k + 1 - i}{2} \right) + D \log 2 - \log|\bm{\Psi}_k| \right)
					- \frac{D}{2} \log(2\pi) \\
					&= - \frac{\nu_k}{2} (\mathbf{x} - \mathbf{m}_k)^\top \bm{\Psi}_k^{-1} (\mathbf{x} - \mathbf{m}_k)
					- \frac{D}{2} s_k^{-1} + \frac{1}{2} \sum_{i = 1}^D \psi\left( \frac{\nu_k + 1 - i}{2} \right) + \frac{1}{2} \log|\bm{\Psi}_k^{-1}|
					- \frac{D}{2} \log\pi.
				\end{align}

	\bibliographystyle{abbrvnat}
	\bibliography{references}